\documentclass[journal]{IEEEtran}

\usepackage{amsmath,amssymb,amsfonts}
\usepackage{mathrsfs}
\usepackage{bm}
\usepackage{cite}
\usepackage{algorithm}
\usepackage{algorithmic}
\usepackage{graphicx}
\usepackage{dblfloatfix}
\usepackage{subcaption}
\usepackage{lipsum}
\usepackage{enumerate}
\usepackage{amsthm}
\usepackage{color}
\usepackage{url}
\usepackage{booktabs}
\usepackage{svg}

\newtheorem{definition}{Definition}
\newtheorem{lemma}{Lemma}
\newtheorem{theorem}{Theorem}

\allowdisplaybreaks[4]

\usepackage{setspace}
\usepackage{threeparttable}
 
\makeatletter
\newcommand*{\rom}[1]{\expandafter\@slowromancap\romannumeral #1@}
\makeatother

\usepackage{accents}
\makeatletter
\def\ubar{\underaccent{{\cc@style\underline{\mskip8mu}}}}
\makeatother

\ifCLASSOPTIONcompsoc
  \usepackage[caption=false,font=normalsize,labelfont=sf,textfont=sf]{subfig}
\else
  \usepackage[caption=false,font=footnotesize]{subfig}
\fi

\begin{document}

\captionsetup[figure]{name={Fig.}, labelsep=period}

\title{Policy Bifurcation in Safe Reinforcement Learning}

\author{Wenjun~Zou$^{1}$,
        Yao~Lyu$^{1}$,
        Jie~Li$^{1}$,
        Yujie~Yang$^{1}$,
        Shengbo~Eben~Li$^{1*}$,
        Jingliang~Duan$^{2}$,
        Xianyuan~Zhan$^{3,4}$,
        Jingjing~Liu$^{3}$,
        Yaqin~Zhang$^{3}$,
        Keqiang~Li$^{1}$
\thanks{$^*$All correspondence should be sent to Shengbo Eben Li. {\tt\small Email: lishbo@mail.tsinghua.edu.cn}. }%
\thanks{$^1$State Key Laboratory of Intelligent Green Vehicle and Mobility, School of Vehicle and Mobility, Tsinghua University, Beijing, China.}%
\thanks{$^2$School of Mechanical Engineering, University of Science and Technology Beijing, China.}%
\thanks{$^3$Institute for AI Industry Research (AIR), Tsinghua University, Beijing, China.}%
\thanks{$^4$Shanghai Artificial Intelligence Laboratory, Shanghai, China.}%
}

\maketitle
\begin{abstract}
Safe reinforcement learning (RL) offers advanced solutions to constrained optimal control problems. 
Existing studies in safe RL implicitly assume continuity in policy functions, where policies map states to actions in a smooth, uninterrupted manner. However, our research finds that in some scenarios, the feasible policy should be discontinuous or multi-valued, and a continuous policy can inevitably lead to constraint violations. We are the first to identify the generating mechanism of such a phenomenon, and employ topological analysis to rigorously prove the existence of policy bifurcation in safe RL, which corresponds to the contractibility of the reachable tuple. 
Our theorem reveals that in scenarios where the obstacle-free state space is non-simply connected, a feasible policy is required to be bifurcated, meaning its output action needs to change abruptly in response to the varying state. To train such a bifurcated policy, we propose a safe RL algorithm called multimodal policy optimization (MUPO), which utilizes a Gaussian mixture distribution as the policy output. The bifurcated behavior can be achieved by selecting the Gaussian component with the highest mixing coefficient. 
Experiments with vehicle control tasks show that our algorithm successfully learns the bifurcated policy and ensures satisfying safety, while a continuous policy suffers from inevitable constraint violations.
\end{abstract}

Reinforcement learning (RL) has made significant advancements in solving complex optimal control problems (OCPs), such as robotic manipulation \cite{kaufmann2023champion}, navigation \cite{bellemare2020autonomous}, and autonomous driving \cite{guan2022integrated, wurman2022outracing, feng2023dense}. The primary objective of RL is to learn an optimal policy by maximizing the expected total rewards. In many practical control tasks, the policy needs to not only aim for reward maximization but also ensure safety by avoiding undesirable states. In an OCP, safety guarantees are often modeled by a set of hard state constraints, which cannot be violated. The corresponding RL algorithms are called safe RL \cite{altman2021constrained,li2023reinforcement}, which has become an effective tool for solving constrained OCPs.

Existing research in safe RL primarily focuses on integrating constrained optimization methods with RL algorithms. J. Achiam et al. \cite{achiam2017constrained} propose constrained policy optimization (CPO), employing trust region methods to achieve near-constraint satisfaction at each iteration theoretically. Alex Ray et al. \cite{ray2019benchmarking} incorporate the Lagrangian multiplier method into proximal policy optimization (PPO) \cite{schulman2017proximal}, resulting in PPO-Lagrangian (PPO-L), which enables constraint satisfaction and return maximization.
Yu et al. \cite{yu2023safe} and Yang et al. \cite{yang2023synthesizing} tackle the problem by exploiting feasible sets in safe RL, defining them as state space regions where policies can consistently operate without violating safety constraints. Their work enhances the understanding and application of feasible sets in safe RL.

\begin{figure}[!ht]
  \centering
  \begin{subfigure}[b]{0.42\textwidth}
    \includegraphics[width=\textwidth]{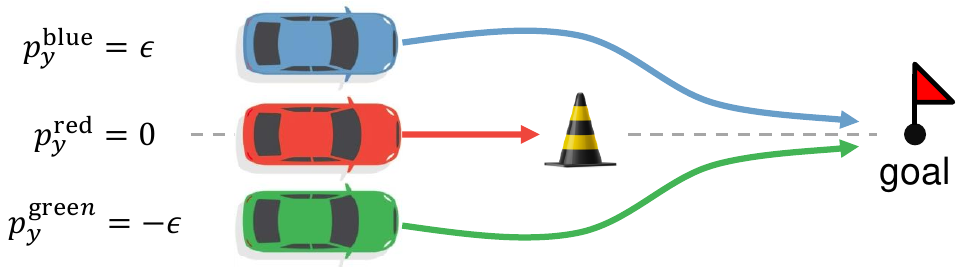}
    \caption{}
    \label{fig:maneuvering_vehicles}
  \end{subfigure}
  \hfill
  \begin{subfigure}[b]{0.40\textwidth}
  \hspace{0.01\textwidth}
    \includegraphics[width=\textwidth]{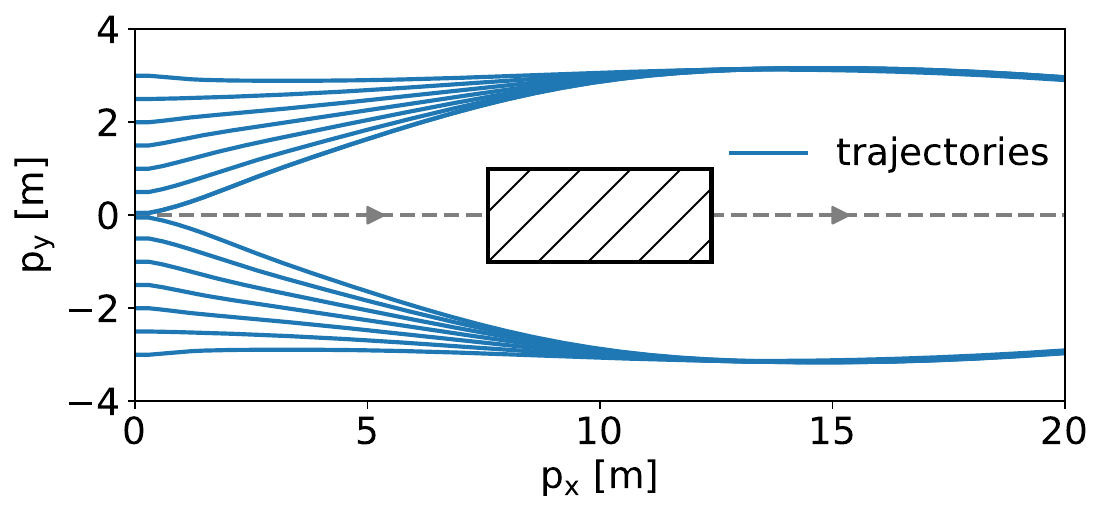}
    \caption{}
    \label{fig:trajectory_clusters}
  \end{subfigure}
  \hfill
  \begin{subfigure}[b]{0.40\textwidth}
    \includegraphics[width=\textwidth]{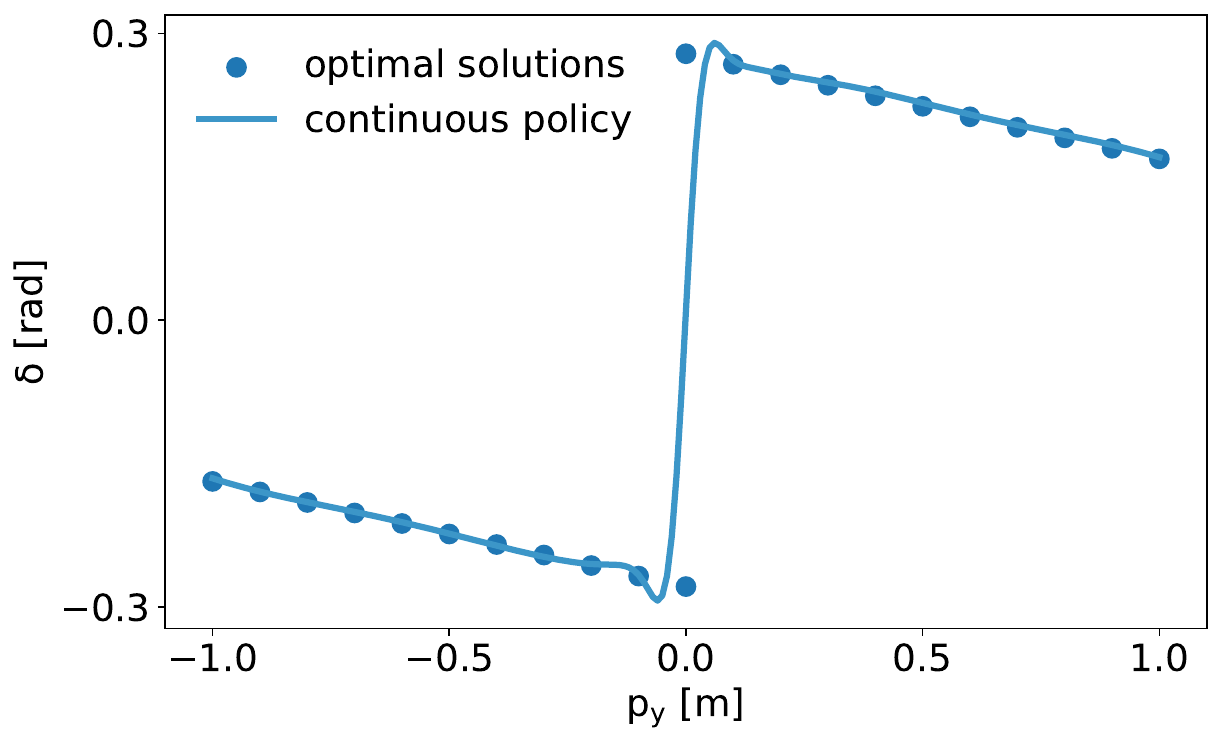}
    \caption{}
    \label{fig:steering_scatter}
  \end{subfigure}
  
  \caption{
  \textbf{Limitations of continuous policies in vehicle control problems.} 
  \textbf{(a)} Illustration of a control problem for autonomous vehicles that encounter an obstacle and must reach a goal region. 
  \textbf{(b)} The trajectories of a vehicle that start from different lateral positions \(p_y\) (in meters) under the optimal policy. There is a bifurcation around \(p_y=0\) where the vehicle steers around the obstacle in different directions.
  \textbf{(c)} Correlation between the front wheel steering angle \(\delta\) (in radians) and lateral position \(p_y\) for an autonomous vehicle starts at a fixed longitudinal position \(p_x\) and speed. The dots show optimal steering responses, with jumps for varying lateral positions at \(p_y=0\). The curve denotes a continuous policy that fails to initiate necessary avoidance maneuvers when starting from the road's center.}
  \label{fig:control_steering_dynamics}
\end{figure}

In terms of policy structure, existing methods apply deep neural networks to approximate policies \cite{achiam2017constrained,ray2019benchmarking,yu2023safe,yang2023synthesizing,ma2022joint,yu2022reachability}. For tasks involving continuous state and action spaces, the inherent continuity of neural networks results in policies that are also continuous, mapping system states to actions smoothly. However, existing research overlooks a serious issue: in many cases, no feasible continuous policy solution may exist for constrained OCPs. In such situations, any continuous policy learned through any safe RL methods will inevitably violate the constraints for some initial states, which is extremely detrimental in practical applications. 

Considering the scenario shown in Fig.~\ref{fig:maneuvering_vehicles}, a safe RL algorithm aims to learn a policy that ensures the ego vehicle reaches the goal without colliding with the obstacle, which is in the center of the road. The ego vehicle starts initially at different positions: blue on the left, red in the center, and green on the right side of the road. The blue vehicle should learn to detour to the left, while the green vehicle should move to the right. As observed, a challenge arises for the red vehicle in the center, as shown in Fig.~\ref{fig:steering_scatter}. If a continuous policy is adopted, the red vehicle may perform an interpolated action based on the actions taken by the blue and green vehicles, potentially leading to a direct collision with the obstacle.

This issue is closely related to multimodal state-action distributions, each of which corresponds to a local optimum of the non-convex OCP. The problem of action multimodality has been noted and studied in imitation learning \cite{florence2022implicit} and offline RL \cite{osa2023offline, ajay2022conditional, lu2023contrastive}, where data sets are derived from different experts. Applying a continuous policy in these contexts can result in interpolation between different modes, which is often low-value and risky. This issue has received litter attention in online RL under the assumption that agents can explore and learn high-value modes while avoiding low-value interpolations. However, as we demonstrate in Fig.~\ref{fig:maneuvering_vehicles}, a continuous policy can result in undesirable intermediate behaviors and lead to constraint violation regardless of the algorithm used. In such scenarios, feasible policies should be bifurcated, where ``bifurcated'' signifies that the policy's actions change abruptly in response to states. We term this phenomenon ``policy bifurcation''. Revealing the underlying mechanisms of this phenomenon remains an unexplored area in current research.

To the best of our knowledge, this work is the first to identify the policy bifurcation mechanism. We established a set of sufficient conditions under which there is no feasible continuous policy for constrained OCPs. These conditions are commonly met in the problem formulations of many robotic and autonomous driving tasks, revealing an overlooked flaw of the policy structure in existing safe RL research.

In this work, we analyze constrained OCPs from a topological perspective. We introduce topological concepts such as paths, loops, and homotopy to define the critical property of contractibility, which plays a pivotal role in our analysis. Our findings reveal that for constrained OCPs with non-simply connected constraints, the set of reachable states and times---termed the reachable tuple---corresponding to a feasible continuous policy must exhibit this contractibility. This requirement often leads to suboptimal policies. Furthermore, when the initial state set is inherently noncontractible, a common occurrence, the reachable tuple cannot be contractible, indicating the impossibility of a continuous policy to be feasible. This highlights the necessity of adopting bifurcated policies in safe RL, especially for tasks with complex safety constraints.

As a solution to the problem revealed by our theoretical findings, we propose the multimodal policy optimization (MUPO) algorithm. It utilizes a Gaussian mixture distribution to realize a bifurcated policy structure. Additionally, it incorporates spectral normalization and forward Kullback-Leibler (KL) divergence, enabling the policy to better capture multimodal action distributions, thereby facilitating the accurate learning of bifurcated policies. Our experimental results demonstrate the ability of MUPO to efficiently learn bifurcated policies that ensure safety in challenging vehicle control tasks.

\subsection*{Problem setting}
We focus on the constrained OCP within a continuous dynamical system, where the objective is to maximize the expected total reward while satisfying safety constraints and reaching the goal. The system is characterized by a continuous state space \( \mathcal{X} \) and an action space \( \mathcal{U} \), with continuous dynamics \( \dot{x}(t) = f(x(t), u(t)) \), where \( x(t) \in \mathcal{X} \) and \( u(t) \in \mathcal{U} \). The control action \( u(t) \) is determined by a policy \( \pi_\theta \) parameterized by \( \theta \), such that \( u(t) = \pi_\theta(x(t)) \). Let \( \mathrm{X}_{\mathrm{init}} \) be the set of initial states from which the system can start, and let \( d_{\mathrm{init}} \) be the probability distribution over \( \mathrm{X}_{\mathrm{init}} \). The safety constraints are denoted by the sub-zero level set of a time-invariant function \( h: \mathcal{X} \rightarrow \mathbb{R} \), defining the constrained set \( \mathrm{X}_{\mathrm{cstr}} = \{ x \in \mathcal{X} \mid h(x) \leq 0 \} \). Conversely, the region where \( h(x) > 0 \) corresponds to the violation region \( \mathrm{X}_{\mathrm{viol}} \), denoting the states that violate the safety constraints. The goal state set \( \mathrm{X}_{\mathrm{goal}} \) denotes the desired outcomes that the system aims to achieve within the time horizon \( T \). The formal statement of the constrained OCP is:

\begin{equation}
\begin{aligned}
& \max_{\theta} \mathbb{E}_{x(0) \sim d_{\mathrm{init}}} \left\{ \int_{0}^{T} \gamma^t r(x(t), u(t)) \, \mathrm{d}t \right\} \\
& \mathrm{s.t.} \quad \begin{aligned}[t]
  & \dot{x}(t) = f(x(t), u(t)), \\
  & u(t) = \pi_\theta(x(t)), \\
  & h(x(t)) \leq 0, \\
  & \forall t \in [0, T], x(0) \in \mathrm{X}_{\mathrm{init}}.
\end{aligned}
\end{aligned}
\label{eq:optimization_problem}
\end{equation}

Safe RL offers a promising approach to address the constrained OCP by learning a policy that improves performance while ensuring safety. In standard safe RL, the policy \( \pi_\theta \) is often parameterized by neural networks. Such a representation of policy usually leads to a continuous policy, which is popular in modern RL algorithms \cite{schulman2017proximal,haarnoja2018soft,hwangbo2019learning,wurman2022outracing}. However, the continuous policy is far from enough for real-world applications, as it will cause unavoidable violations of safety constraints. As shown in Fig.~\ref{fig:trajectory_clusters} and Fig.~\ref{fig:steering_scatter}, extensive scenarios require us to design a bifurcated policy to ensure safety. The mechanism of policy bifurcation for the constrained OCP remains an open problem, which is the main focus of this work.

\section*{Results}

\subsection*{Suboptimality and infeasibility of continuous policies}

\begin{figure*}[!t]
  \centering
  \begin{subfigure}[t]{0.48\textwidth}
    \centering
    \includegraphics[width=0.8\linewidth]{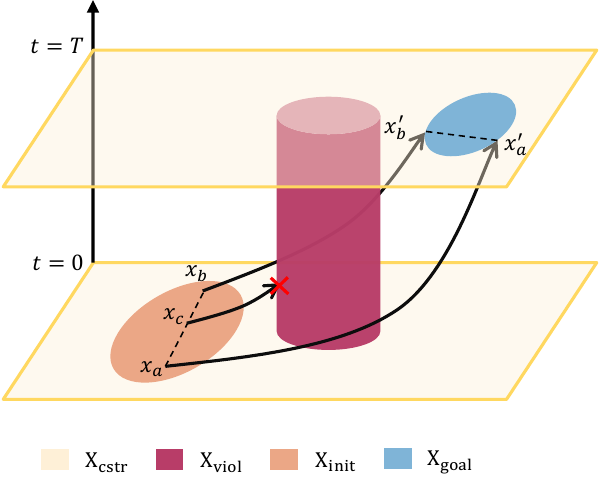}
    \caption{}
    \label{fig:theorem1_1}
  \end{subfigure}
  \hfill
  \begin{subfigure}[t]{0.48\textwidth}
    \centering
    \includegraphics[width=0.8\linewidth]{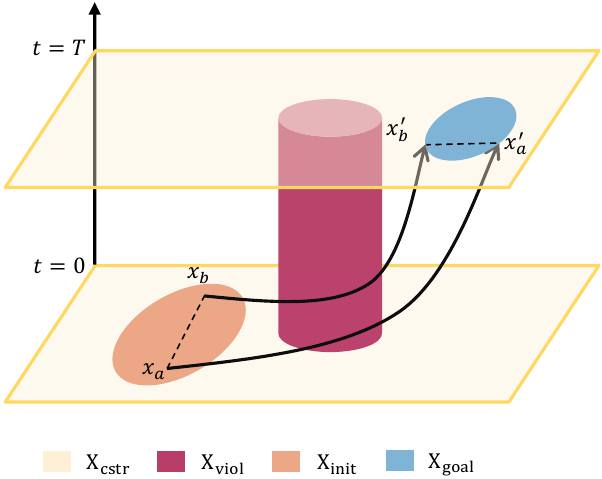}
    \caption{}
    \label{fig:theorem1_2}
  \end{subfigure}
  \caption{\textbf{Suboptimality of continuous policies.} \textbf{(a)} Illustration of a constrained OCP in a 2D state space, extended with a time dimension. Due to the fact that the safety constraints are active at all times, the violation region \( \mathrm{X}_{\mathrm{viol}} \) is depicted as a cylinder that extends along the time dimension. For the open-loop optimal solution, trajectories from initial states \( x_\mathrm{a} \) and \( x_\mathrm{b} \) to goal states \( x_\mathrm{a}' \) and \( x_\mathrm{b}' \) bypass the obstacle on different sides. The loop \( \ell_{aa'b'ba} \) cannot be continuously contracted to a point within \( \mathrm{X}_{\mathrm{cstr}} \) without passing through \( \mathrm{X}_{\mathrm{viol}} \), due to the non-simply connected property of \( \mathrm{X}_{\mathrm{cstr}} \). According to our theoretical analysis, for any continuous policy acting in this manner, there must exist an initial state \( x_{c} \), from which the trajectory must violate the constraints. \textbf{(b)} Illustration of a feasible continuous policy for the constrained OCP, showing the trajectory that avoids the obstacle from one side to ensure the feasibility of the continuous policy. Continuous policy is forced to take a significantly suboptimal trajectory compared to open-loop optimal solutions, leading to a substantial loss in optimality.}
  \label{fig:theorem1}
\end{figure*}

We introduce two key concepts in our analysis: the \textit{reachable tuple} $\mathcal{R}$, and \textit{contractibility}. The reachable tuple $\mathcal{R}$ denotes the set of all state-time pairs \((x,t)\) that the system can reach within a given time frame \(T\), meaning the system can be in state $x$ at time $t$. Contractibility refers to the property that a set can be continuously transformed into a single point without violating the constraints of the system. It is important to note that this definition differs from the traditional notion of contractibility in topology. When the constrained set $\mathrm{X}_\mathrm{cstr}$ is non-simply connected, its subset could be noncontractible. This noncontractibility further leads to limitations of continuous policies in constrained OCPs, which are presented as follows:

\textbf{Suboptimality of continuous policies:} for a constrained OCP characterized by a Lipschitz continuous dynamic function $f$, if the optimal policy corresponds to a reachable tuple $\mathcal{R}$ that is noncontractible, then the optimal policy cannot be achieved by continuous policy. Fig.~\ref{fig:theorem1} provides an illustrative example of this theorem.

\begin{figure}[!ht]
    \centering
    \includegraphics[width=0.38\textwidth]{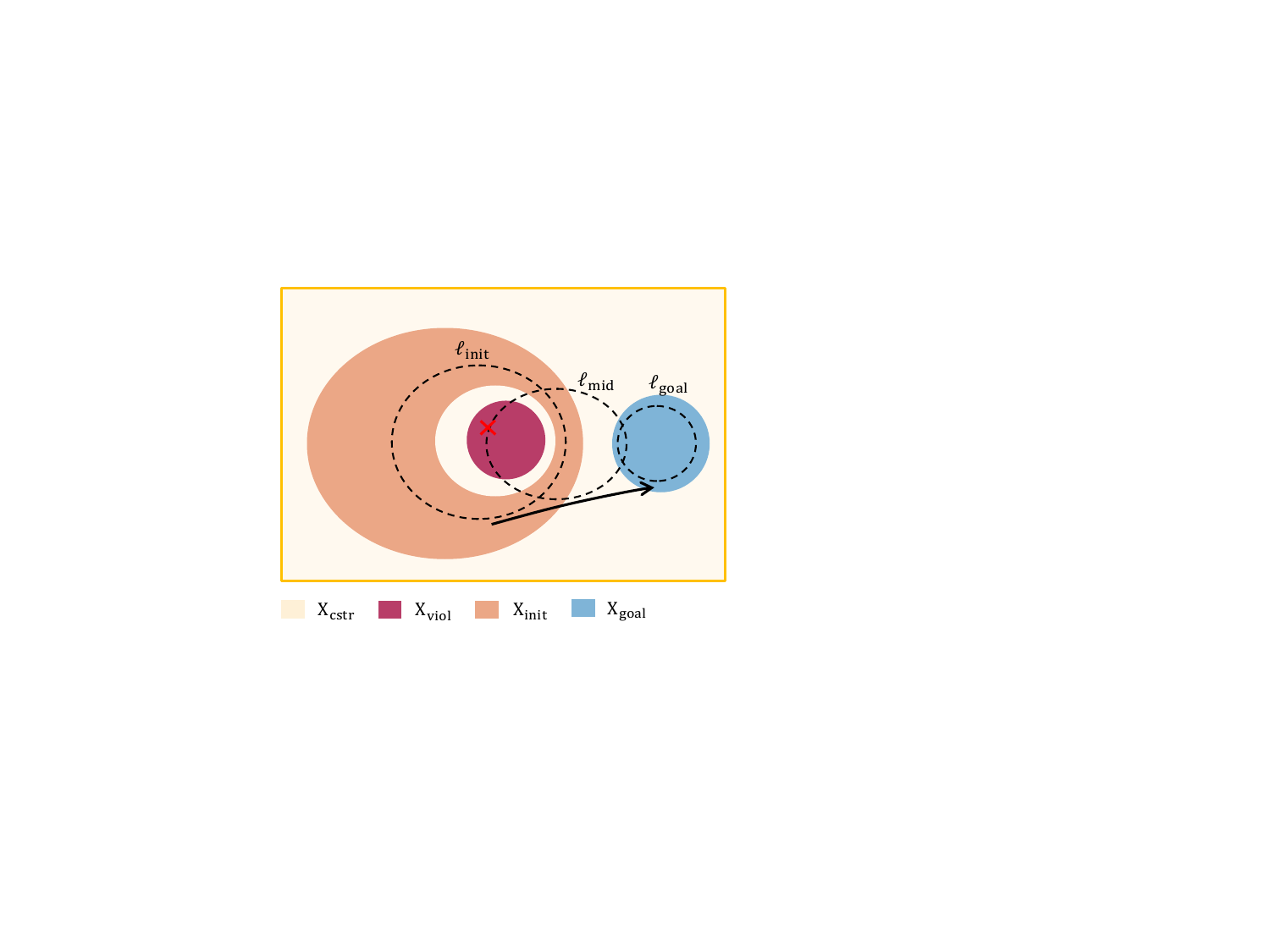}
    \caption{\textbf{Infeasibility of continuous policies.} This figure illustrates a constrained OCP within a 2D state space. The initial state set $\mathrm{X}_{\mathrm{init}}$ is noncontractible; for instance, a closed curve $\ell_{\mathrm{init}}$ within $\mathrm{X}_{\mathrm{init}}$ encircles the violation region $\mathrm{X}_{\mathrm{viol}}$. Under a continuous policy, all points on $\ell_{\mathrm{init}}$ after the same time period form a new closed curve, such as $\ell_{\mathrm{mid}}$ in the figure. However, there exists no continuous deformation in topology that can transform $\ell_{\mathrm{init}}$ into $\ell_{\mathrm{goal}}$ without intersecting $\mathrm{X}_{\mathrm{viol}}$, thus demonstrating the infeasibility of the continuous policy.}
    \label{fig:theorem2}
\end{figure}

\textbf{Infeasibility of continuous policies:} if the initial state set $\mathrm{X}_\mathrm{init}$ is noncontractible, and the goal state set $\mathrm{X}_\mathrm{goal}$ is contractible, then a constrained OCP has no feasible continuous policy. Fig.~\ref{fig:theorem2} provides an illustrative example of this theorem.


These theorems highlight the inherent limitations of continuous policies in constrained OCPs. Importantly, these limitations are intrinsic to the continuous policies and are not dependent on the specific learning algorithm. As long as the policy is continuous, these challenges in achieving optimality and feasibility under non-simply connected constraints will persist.

\subsection*{Experimental validation for learning bifurcated policy}
\label{sec.experiments}

\begin{figure*}[!t]
  \centering
  \begin{subfigure}[t]{0.23\textwidth}
    \includegraphics[width=\textwidth]{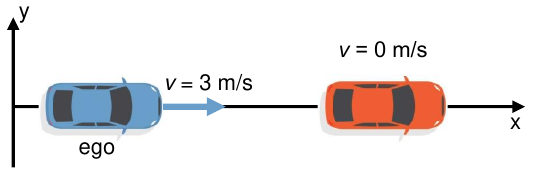}
    \caption{}
    \label{fig:task_bypass}
  \end{subfigure}
  \begin{subfigure}[t]{0.23\textwidth}
    \includegraphics[width=\textwidth]{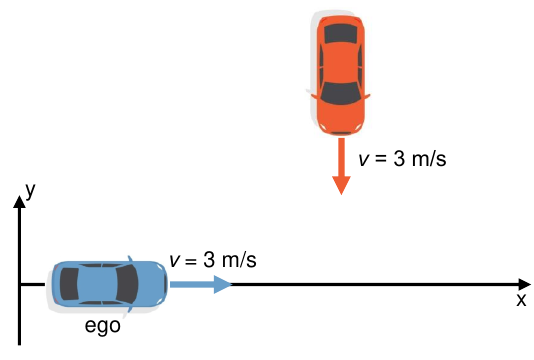}
    \caption{}
    \label{fig:task_encounter}
  \end{subfigure}
  \begin{subfigure}[t]{0.24\textwidth}
    \includegraphics[width=\linewidth]{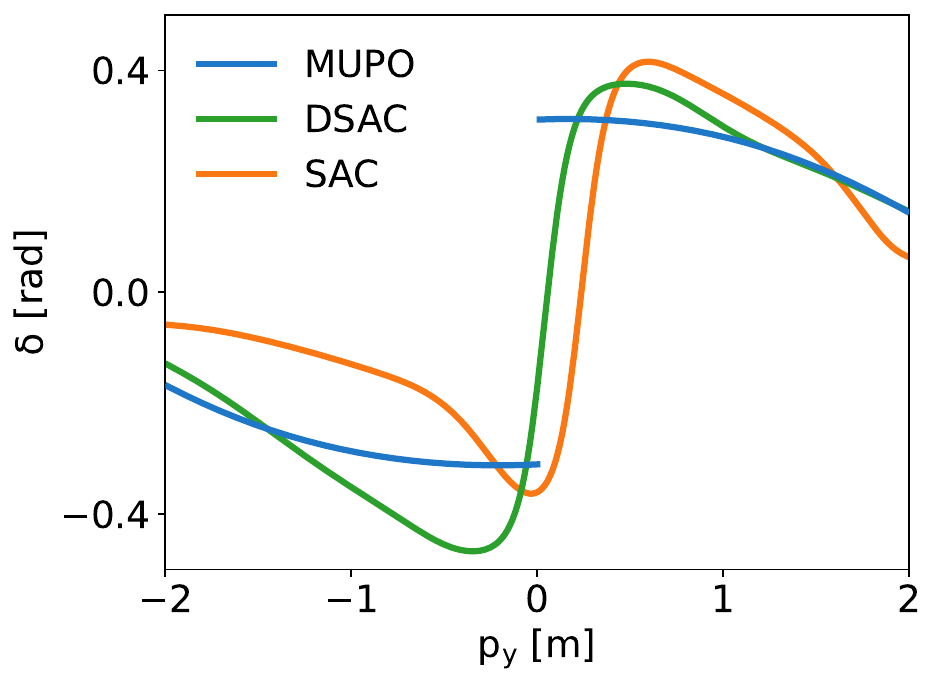}
    \caption{}
    \label{fig:bypass_steer}
  \end{subfigure}
  \begin{subfigure}[t]{0.24\textwidth}
    \includegraphics[width=\linewidth]{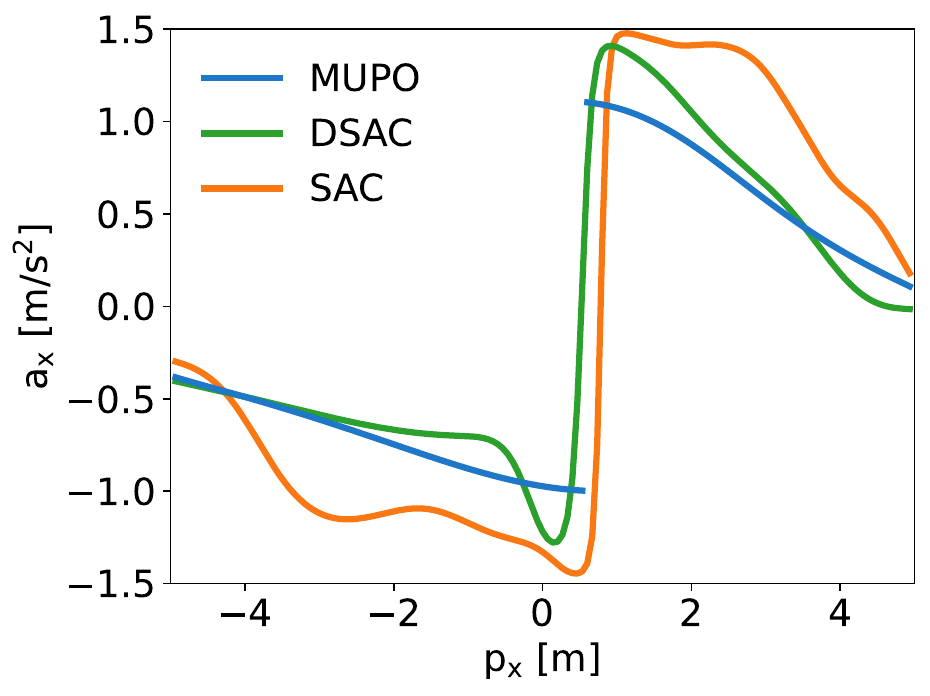}
    \caption{}
    \label{fig:encounter_acc}
  \end{subfigure}
  \begin{subfigure}[t]{0.49\textwidth}
    \includegraphics[width=\linewidth]{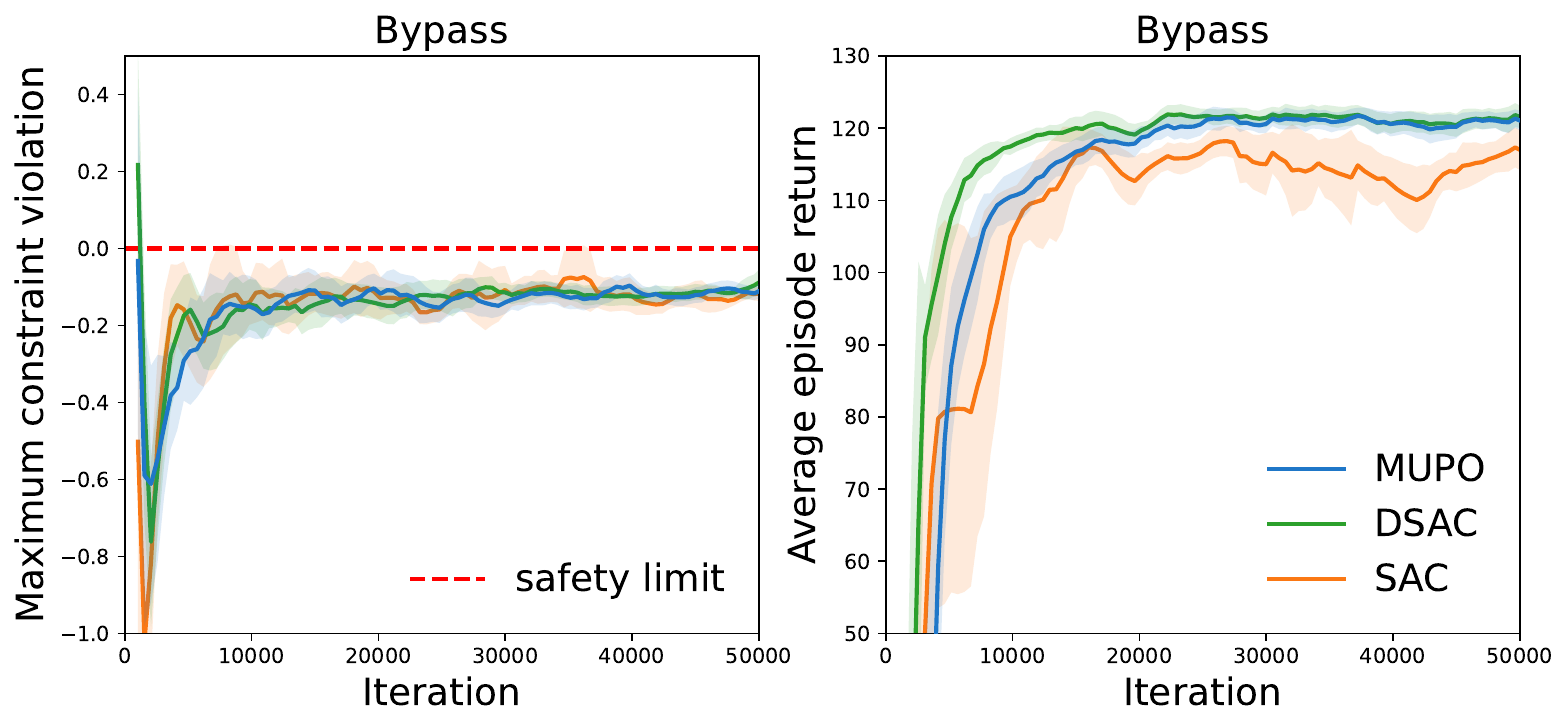}
    \caption{}
    \label{fig:bypass_combined}
  \end{subfigure}
  \hfill
  \begin{subfigure}[t]{0.49\textwidth}
    \includegraphics[width=\linewidth]{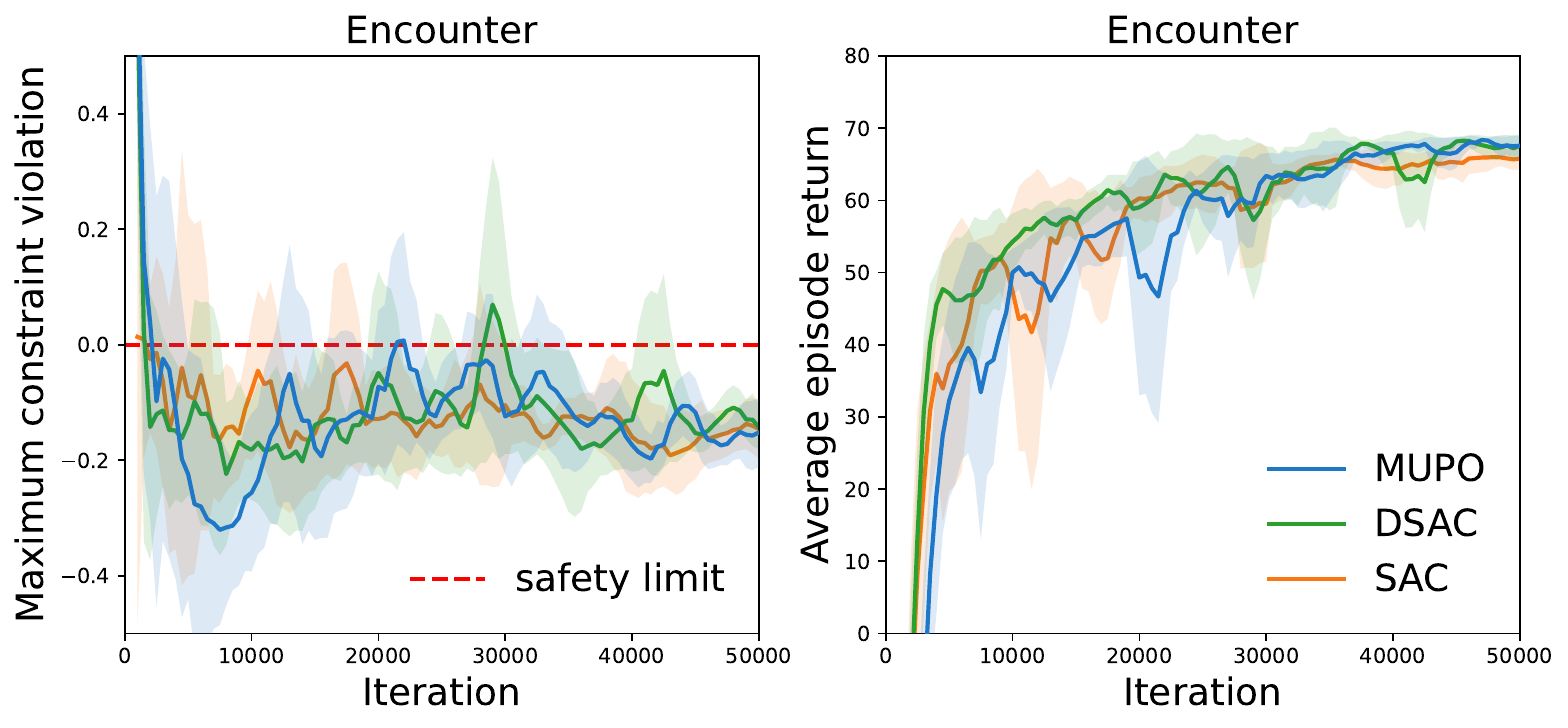}
    \caption{}
    \label{fig:encounter_combined}
  \end{subfigure}
  
  \begin{subfigure}[t]{0.48\textwidth}
    \includegraphics[width=\linewidth]{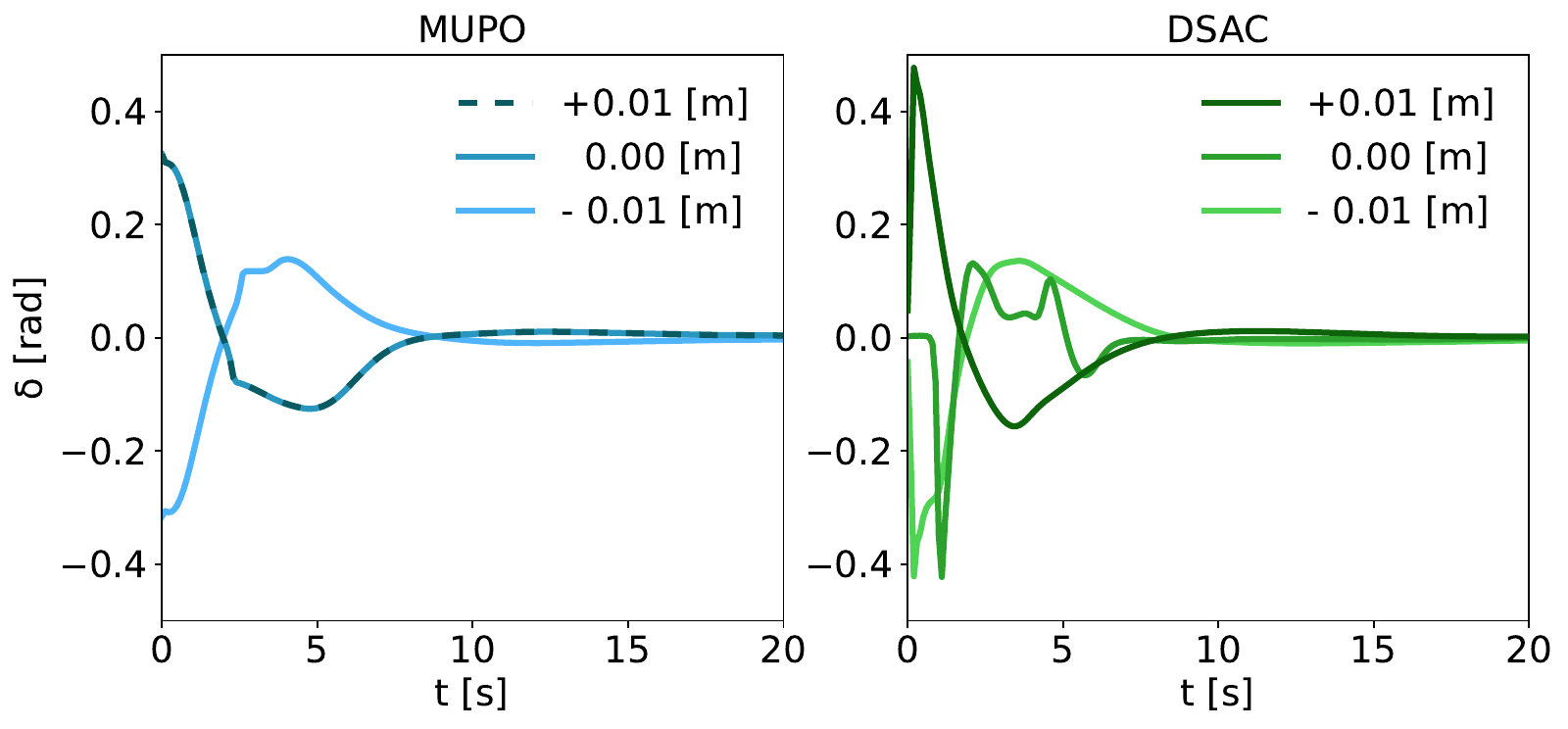}
    \caption{}
    \label{fig:control_output}
  \end{subfigure}
  \hfill
  \begin{subfigure}[t]{0.48\textwidth}
    \includegraphics[width=\linewidth]{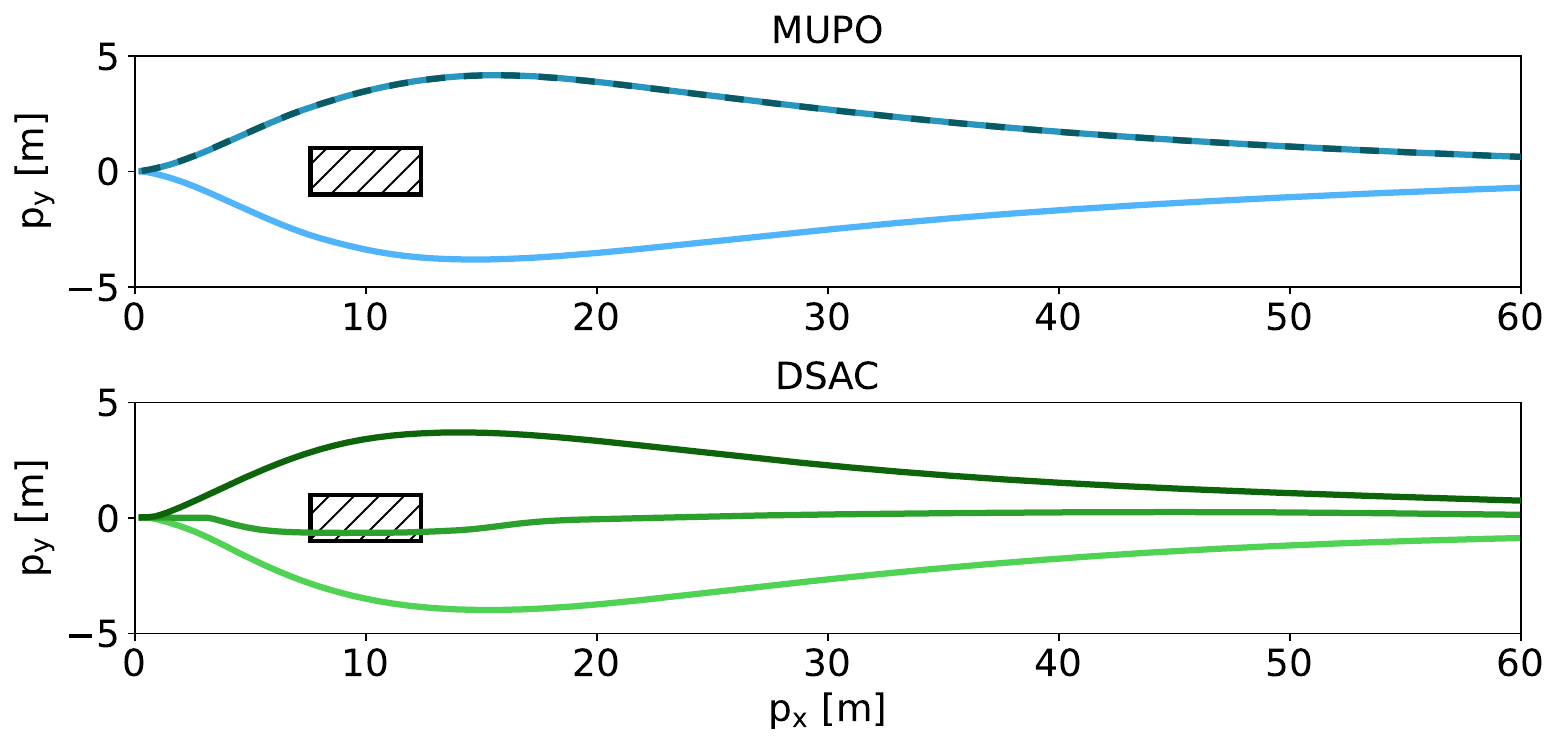}
    \caption{}
    \label{fig:control_effect}
  \end{subfigure}

  \caption{\textbf{Experimental results and visualization.} \textbf{(a)} Bypass task illustration. \textbf{(b)} Encounter task illustration. \textbf{(c)} Open-loop control for the Bypass task. \textbf{(d)} Open-loop control for the Encounter task. \textbf{(e)} Training curves for the Bypass task. \textbf{(f)} Training curves for the Encounter task. \textbf{(g)} Front-wheel steering angle \(\delta\) changes over time for the Bypass task where the vehicle starts from different lateral positions \(p_y\). \textbf{(h)} Trajectories for the Bypass task where the vehicle starts from different lateral positions \(p_y\).}
  \label{fig:combined_results}
\end{figure*}

To address the limitations of continuous policies, we construct a bifurcated policy that outputs the parameters of a Gaussian mixture distribution. By selecting the mean of the Gaussian component with the highest gate probability as the action, this approach allows for discontinuous changes in the action with respect to changes in the state. We propose the MUPO algorithm, developed based on maximum entropy RL \cite{haarnoja2018soft}, to train this bifurcated policy. It utilizes the forward KL divergence to enhance its ability to learn multimodal distributions, avoiding convergence to suboptimal solutions. Our experimental validation demonstrates that the bifurcated policies learned by the MUPO algorithm can effectively ensure safety and optimality in vehicle control tasks. In contrast, continuous control policies are shown to lead to constraint violations in these scenarios.

\subsubsection{Simulation experiments}
\label{subsec.simulation}

We build two experimental tasks with a three-degree-of-freedom vehicle model \cite{ge2021numerically}. The primary objective of these tasks is to effectively maneuver the ego vehicle along a linear trajectory while adeptly circumventing any potential obstacles. The first task, referred to as ``Bypass", involves a stationary obstacle on the reference path. The objective of the ego vehicle is to bypass this obstacle and then return to the reference path. The second task, referred to as ``Encounter", involves dealing with a vehicle approaching at a constant speed from the left side. The goal for the ego vehicle in this scenario is to maintain on the reference path while ensuring there is no collision with the approaching vehicle. 

The baselines for comparison are distributional soft actor-critic (DSAC) \cite{duan2023dsac} and soft actor-critic (SAC) \cite{haarnoja2018soft} with continuous policies. Detailed hyperparameters can be found in the supplementary materials. First, we evaluate the algorithms with two metrics: maximum constraint violation and average episode return. The training curves are shown in Fig.~\ref{fig:bypass_combined} and Fig.~\ref{fig:encounter_combined}. The results demonstrate that both MUPO and DSAC achieve comparable high average episode returns, surpassing the performance of SAC, while seemingly satisfying the safety constraints under the current evaluation setup. However, it is important to note that the evaluation based on a limited number of random initializations cannot fully reflect the safety of the learned policies. Our next experiment will further show that only bifurcated policy can guarantee safety, while continuous policy cannot.

\begin{figure*}[!t]
  \centering
  \begin{subfigure}[t]{0.49\textwidth}
  \hspace*{0.11\textwidth}
    \includegraphics[width=0.85\linewidth]{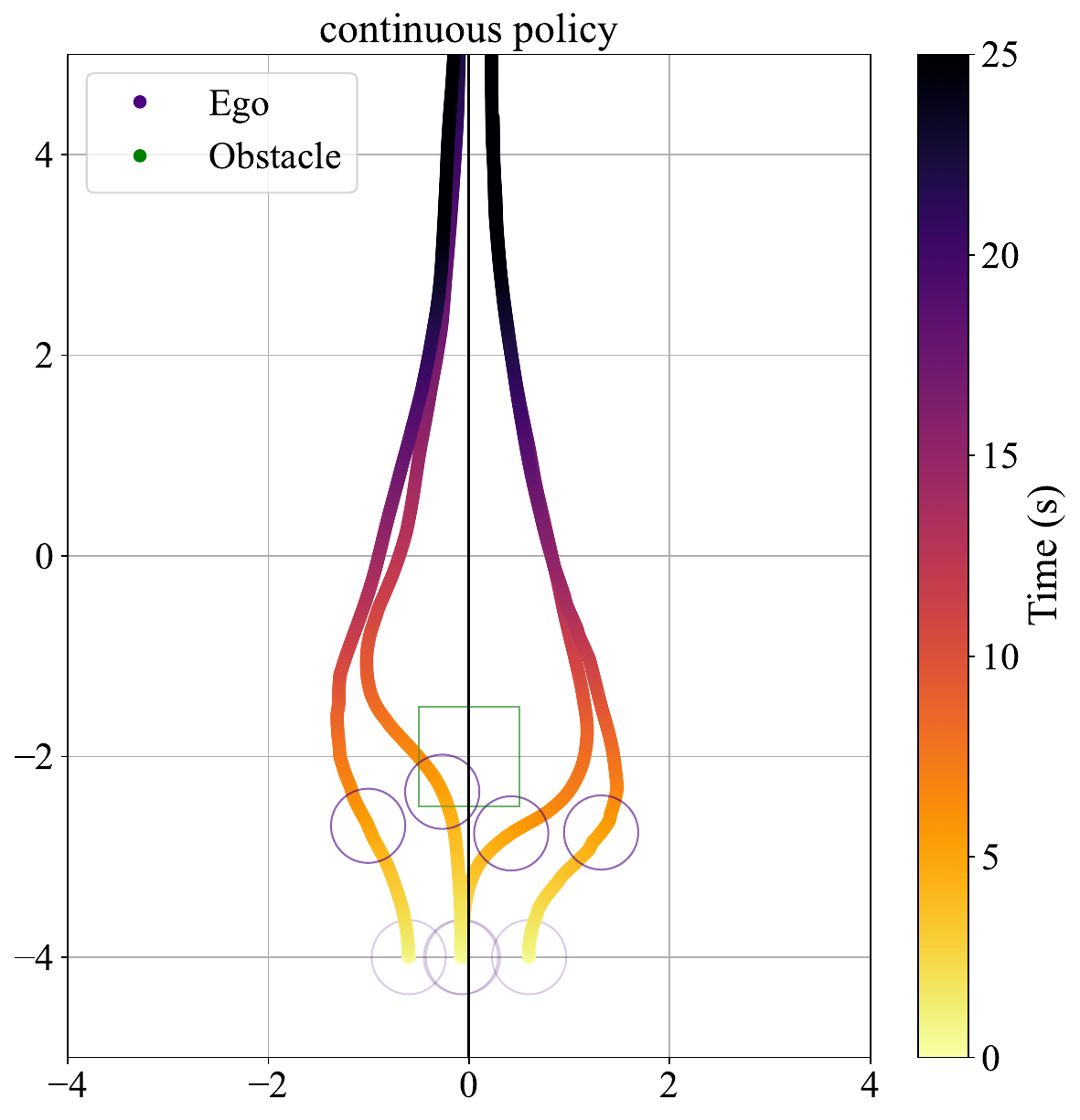}
    \caption{}
    \label{fig:ugv_conti}
  \end{subfigure}
  \hfill
  \begin{subfigure}[t]{0.49\textwidth}
  \hspace*{0.11\textwidth}
    \includegraphics[width=0.85\linewidth]{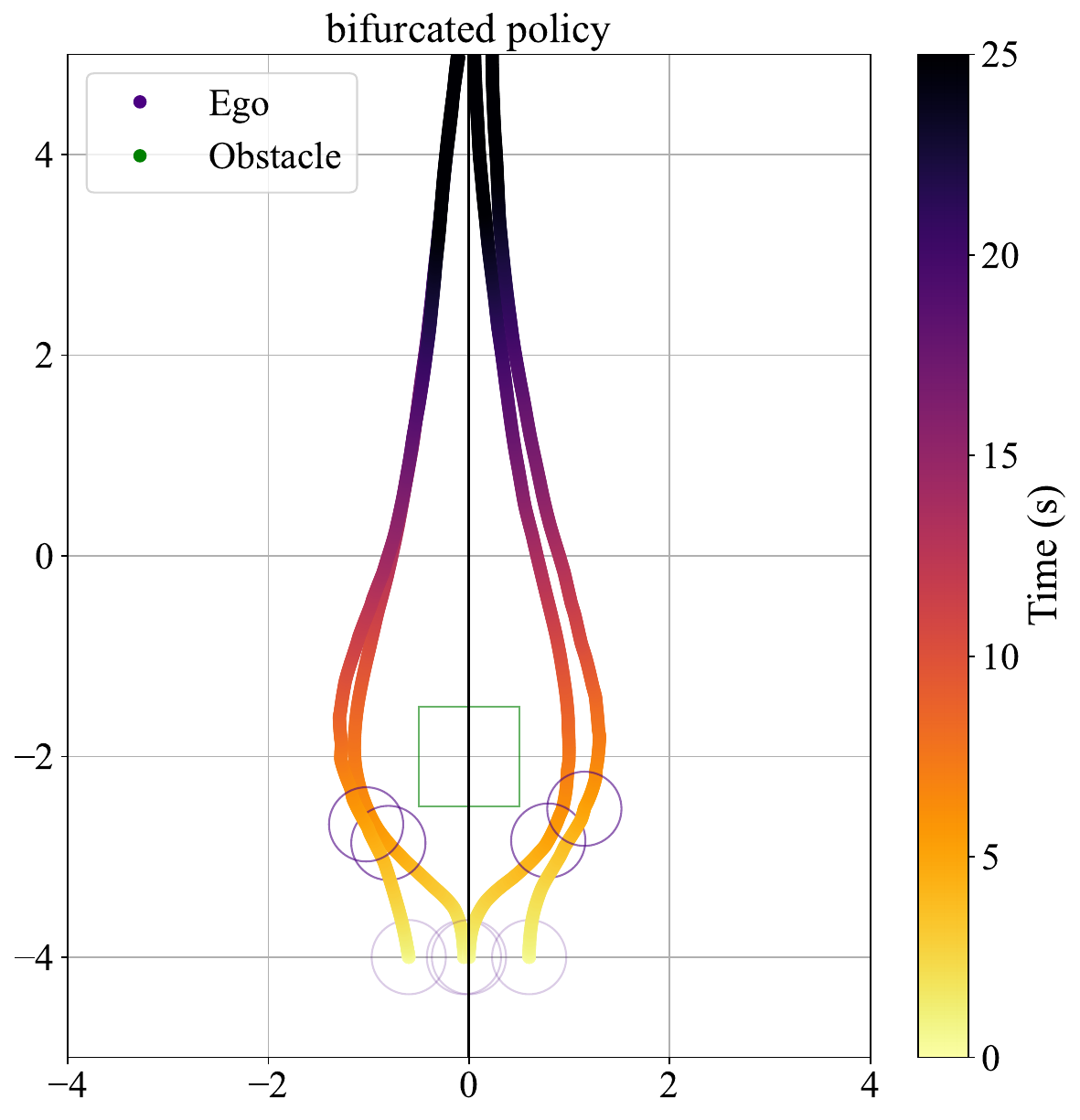}
    \caption{}
    \label{fig:ugv_bifur}
  \end{subfigure}
  \begin{subfigure}[t]{0.49\textwidth}
    \includegraphics[width=0.9\linewidth]{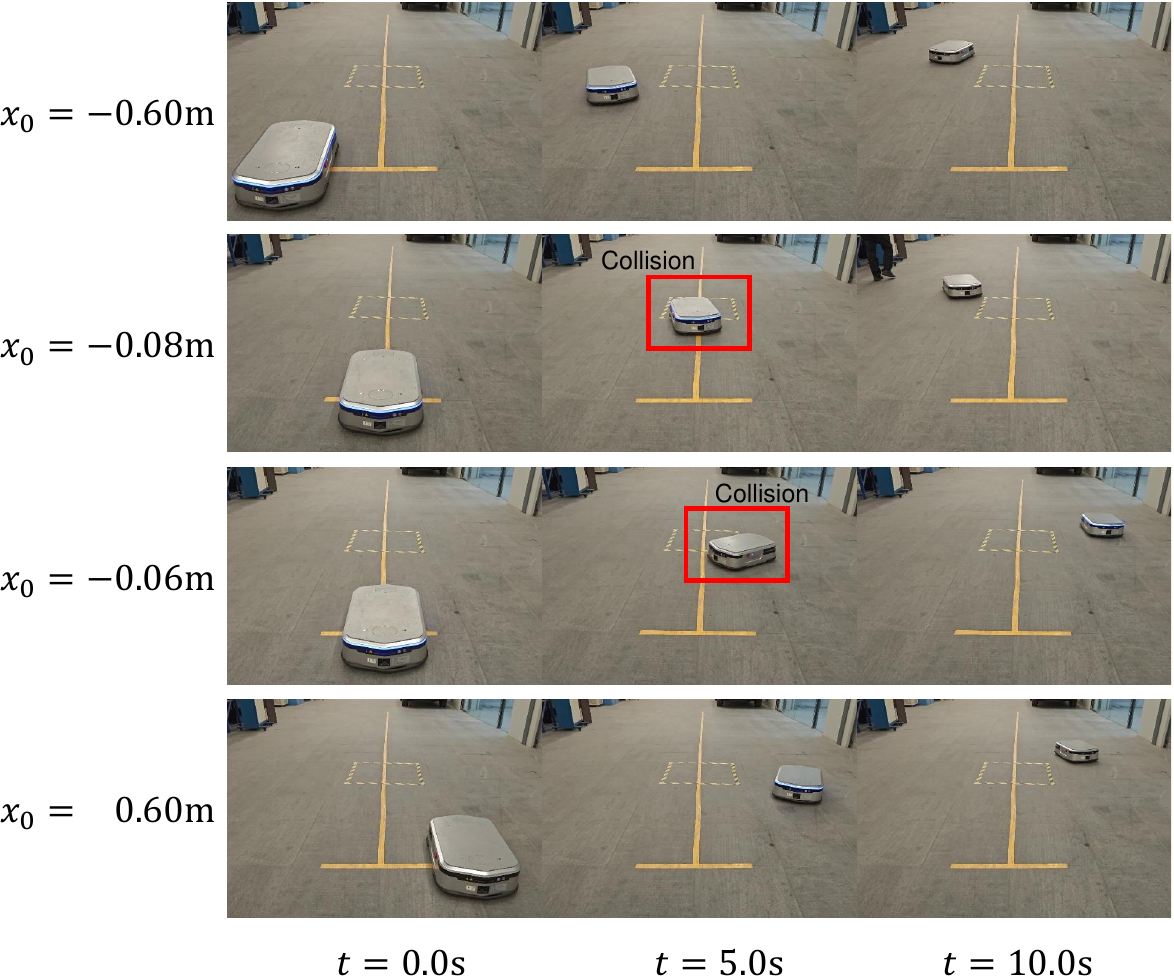}
    \caption{}
    \label{fig:ugv_photo_conti}
  \end{subfigure}
  \hfill
  \begin{subfigure}[t]{0.49\textwidth}
    \includegraphics[width=0.9\linewidth]{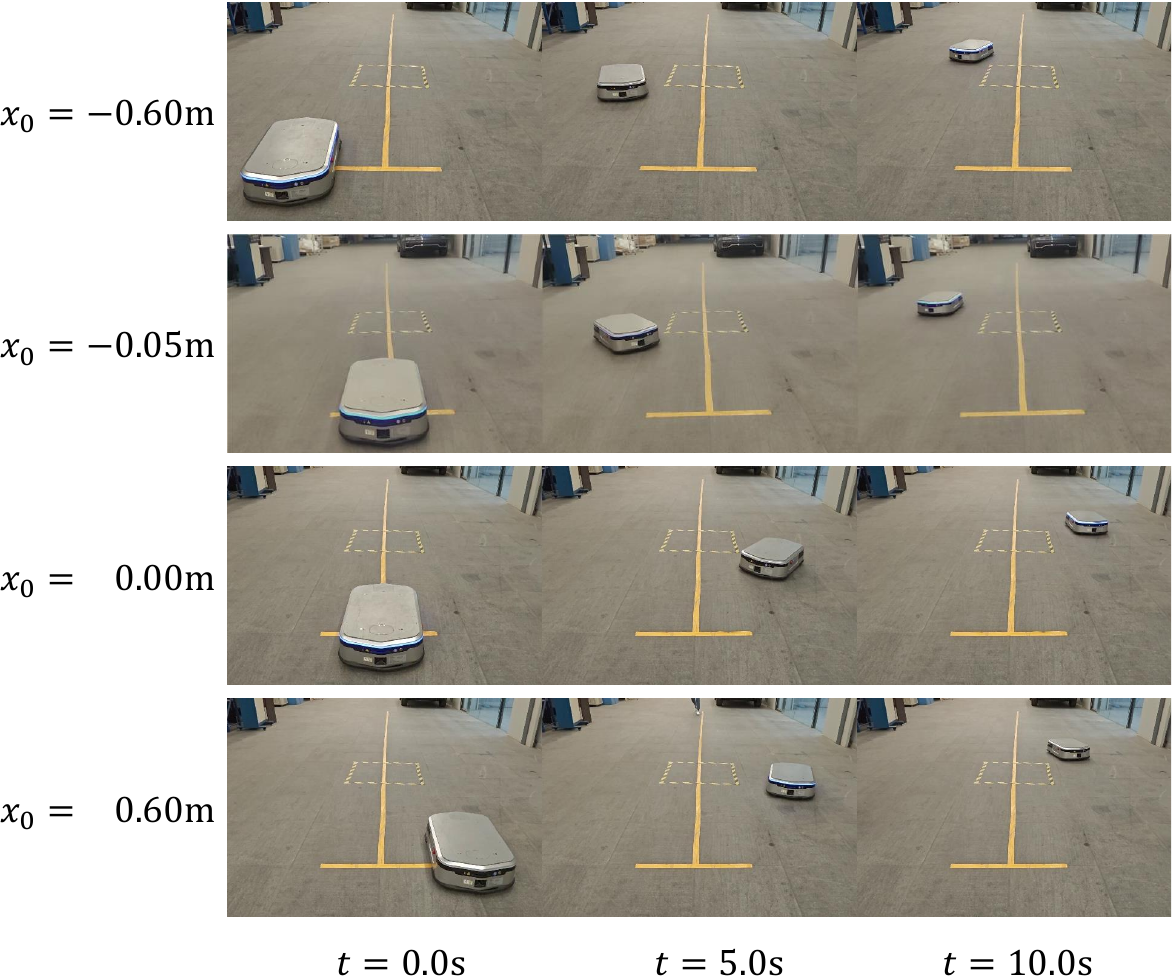}
    \caption{}
    \label{fig:ugv_photo_bifur}
  \end{subfigure}
  \caption{\textbf{Comparison of autonomous vehicle trajectories under continuous versus bifurcated policies in real-world execution.} \textbf{(a)} Trajectory visualization of autonomous driving with continuous policy. \textbf{(b)} Trajectory visualization of autonomous driving with bifurcated policy. \textbf{(c)} Snapshot of autonomous vehicle executing continuous policy. \textbf{(d)} Snapshot of autonomous vehicle executing bifurcated policy.}
\label{fig:ugv_comparison}
\end{figure*}

We plot the action variations against initial states for both the bifurcated policies of MUPO and the continuous policies of DSAC and SAC. In the Bypass task, Fig.~\ref{fig:bypass_steer} shows how the front-wheel steering angle \( \delta \) changes with the initial lateral position \( p_y \). All the policies tend to steer left (positive front-wheel angle \( \delta \)) when initially positioned on the left side of the road, and steer right on the right side. However, near the road center, continuous policies interpolate between left and right steering—going straight, while bifurcated policies can exhibit discontinuous transitions. In the Encounter task, Fig.~\ref{fig:encounter_acc} shows how the longitudinal acceleration \( a_x \) changes with the initial longitudinal position \( p_x \). All the policies accelerate when initially ahead, and decelerate when behind, to pass or yield, respectively. Again, only bifurcated policies exhibit discontinuous behavioral transitions. In both tasks, the MUPO learns bifurcated policies that exhibit discontinuous transitions in actions as the state changes, which helps avoid constraint violations.

In our closed-loop simulation, we analyze the behavior of the ego vehicle initialized at three specific lateral positions: +0.01m, 0.00m, and -0.01m. The control curves of the front-wheel steering angle are displayed in Fig.~\ref{fig:control_output}, and the trajectories of the ego vehicle are shown in Fig.~\ref{fig:control_effect}, with the obstacle represented by a black rectangle. The policy of MUPO demonstrates a bifurcation characteristic, where infinitesimal variations in the initial state induce significant shifts in behavioral modes. On the other hand, the DSAC adopts a continuous policy, adjusting behavior smoothly in response to initial state changes. However, between left and right bypass maneuvers around an obstacle, there exists no continuous transition that satisfies constraints. Consequently, employing the continuous policy inevitably leads to constraint violations. It is worth noting that the initial states leading to constraint violations constitute a minute fraction of the overall state space, as evidenced by position deviation presented at a sub-centimeter scale. This explains why previous studies have not identified this issue: the probability of constraint violations is extremely low when initial states are randomly selected. However, the non-zero probability of such violations occurring cannot be overlooked. In scenarios where safety is critical, the adoption of bifurcated policies is essential.

\subsubsection{Real-world experiments}
\label{subsec.realworld}
We explore the potential for constraint violations under continuous policies and evaluate MUPO's capacity to ensure safety in the real world. We utilize the Geekplus M200 mobile robot in our experiments, which is programmed to follow a linear trajectory at 0.3m/s while avoiding a centrally placed square obstacle, as illustrated in Fig.~\ref{fig:ugv_comparison}. The robot is initialized randomly from various positions, as depicted in Fig.~\ref{fig:ugv_photo_conti} and Fig.~\ref{fig:ugv_photo_bifur}. Our observations, as depicted in Fig.~\ref{fig:ugv_conti}, confirm that the robot could avoid the obstacle when initiated from the sides under continuous policy control. However, initiating from a central position led to a collision. In contrast, as shown in Fig.~\ref{fig:ugv_bifur}, the bifurcated policy effectively facilitates obstacle avoidance for the robot, regardless of its initial position. This empirical observation aligns with our theoretical predictions, showcasing that the bifurcation property of the policy is crucial in meeting constraints and approaching the global optimum.

\section*{Discussion}

Our study has revealed the inherent limitations of continuous policies in constrained OCPs. Specifically, our main theoretical results establish that continuous policies fail to achieve optimality and feasibility when the system's constrained set is non-simply connected. If the initial state set is noncontractible while the goal state set is contractible, no feasible continuous policy exists for the constrained OCP. Although our analysis concentrates on these topological conditions, they are not rare exceptions but rather common occurrences in a wide range of real-world control tasks. This underscores the need for alternative approaches in the design of control policies for constrained OCPs.

The implications of this finding are profound for the field of safe RL, challenging the prevailing use of continuous policy functions in mainstream RL approaches. Our study introduces a novel perspective by highlighting the necessity of adopting bifurcated policies to address the topological challenges in constrained OCPs, a consideration that has been largely overlooked in existing literature. The proposed MUPO algorithm represents a significant advancement in this direction, learning a bifurcated policy that ensures safety and optimality in vehicle control tasks. Experimental validation confirms that MUPO's bifurcated policies can ensure safety by exhibiting behavior patterns that abruptly change with the system's state, whereas continuous control policies often result in constraint violations. However, it is important to note that MUPO applies a mixture of Gaussian distributions, which can not approximate arbitrary distributions, leading to potential limitations in performance. Furthermore, while existing methods like diffusion models \cite{yang2023policy,zheng2024safe,chi2023diffusion} can fit arbitrary continuous distributions, they often involve high computational complexity to obtain the most probable actions. Future work will aim to address these challenges to enhance the applicability and effectiveness of the MUPO algorithm.

In conclusion, our study identifies a key limitation in existing RL methods for handling safety control tasks with continuous state and action spaces. This suggests a need to move away from traditional continuous policy functions and towards policy network structures with bifurcation properties. The MUPO algorithm represents a significant advancement in formulating safe and effective control policies for such challenges, with potential applications extending to autonomous driving and other safety-critical domains.

\section*{Methods}

\subsection*{Continuous state transition in control system}

In this work, we focus on constrained OCP with continuous state space, action space, and dynamic functions, as introduced in the Problem setting section. When dealing with such continuous control problems, existing RL methods primarily adopt continuous policy functions, based on the assumption that continuous functions can satisfy both feasibility and optimality. However, this work challenges that assumption. To begin with, such continuous control systems result in continuous state transitions.

\begin{lemma}
\label{lemma.continuous_state_transition}
Given the control system defined by the dynamic function \( f \) and the control policy \( \pi \), we define the continuous-time state transition function as \( F_{\pi}: \mathcal{X} \times \mathbb{R}_{\geq 0} \to \mathcal{X} \). \(F_{\pi}(x_\mathrm{init},t)\), which maps an initial state \( x_\mathrm{init} \) to the subsequent state at a later time \( t \) under the policy \( \pi \). It is determined by solving the initial value problem:
\begin{equation*}
\frac{dx}{dt} = f(x, \pi(x)), \quad x(0) = x_\mathrm{init}.
\end{equation*}

Here, the relationship between \(F_{\pi}\) and the solution of the differential equation is given by:
\begin{equation*}
F_{\pi}(x_{\mathrm{init}}, t) = x(t) |_{x(0) = x_{\mathrm{init}}}.
\end{equation*}

Under the assumption of Lipschitz continuity for \( f \) and \( \pi \), the solution \( F_{\pi}(x,t) \) is guaranteed to exist uniquely and be continuous with respect to both \( x \) and \( t \).
\end{lemma}

\begin{proof}
Given that \( f \) and \( \pi \) are Lipschitz continuous functions, let us define a composite function \( g(x) = f(x, \pi(x)) \) that denotes the right-hand side of the state's differential equation. By the properties of Lipschitz functions, the function \( g \) is also Lipschitz continuous since for any two points \( x_1, x_2 \in \mathcal{X} \), the Lipschitz condition gives:
\begin{equation*}
\begin{aligned}
\| g(x_1) - g(x_2) \| &= \| f(x_1, \pi(x_1)) - f(x_2, \pi(x_2)) \| \\
&\leq L_f(\| x_1 - x_2 \| + \| \pi(x_1) - \pi(x_2) \|) \\
&\leq L_f(1 + L_{\pi})\| x_1 - x_2 \|,
\end{aligned}
\end{equation*}
where \( L_f \) and \( L_{\pi} \) are the Lipschitz constants of \( f \) and \( \pi \), respectively. The Picard-Lindelöf theorem \cite{hartman2002ordinary} asserts that for a Lipschitz continuous function \( g \), there exists a unique solution to the initial value problem \( \dot{x}(t) = g(x(t)) \) for each initial condition \( x(0) \) in \( \mathcal{X} \). Therefore, the state transition function \( F_{\pi} \), which describes the state of the system at any time \( t \), exists uniquely for each initial state \( x \) and is continuous with respect to both \( x \) and \( t \).
\end{proof}


\subsection*{Time-augmented state dynamics}

\begin{definition}[Augmented state]
\label{def:augmented_state}
\textnormal{
To facilitate subsequent proofs in the context of control systems, we consider augmenting the system's state by incorporating time. Specifically, time \( t \) is introduced as an additional dimension to the system state, leading to the augmented state denoted as \( \tilde{x} = (x, t) \). The augmented state space is defined as
\begin{equation*}
\tilde{\mathcal{X}} = \mathcal{X} \times \mathbb{R}_{\geq 0}.
\end{equation*}
}

\textnormal{
The constraint set \( \mathrm{X}_{\mathrm{cstr}} \) is also augmented to align with \( \tilde{\mathcal{X}} \), resulting in the augmented constraint set defined as
\begin{equation*}
\tilde{\mathrm{X}}_{\mathrm{cstr}} = \{ (x, t) \in \tilde{\mathcal{X}} \mid h(x) \leq 0 \}.
\end{equation*}
}

\textnormal{
The augmented initial set \( \tilde{\mathrm{X}}_{\mathrm{init}} \) is defined as
\begin{equation*}
\tilde{\mathrm{X}}_{\mathrm{init}} = \{ (x, 0) \mid x \in \mathrm{X}_{\mathrm{init}} \}.
\end{equation*}
}
\end{definition}

To clearly delineate the components of the augmented state \(\tilde{x}\), we introduce two projection functions denoted by \(\phi\). Specifically, \(\phi_x(\tilde{x}) = x\) extracts the spatial component of the augmented state, whereas \(\phi_t(\tilde{x}) = t\) extracts the temporal component. These functions allow us to refer to and manipulate the individual elements of \(\tilde{x}\) in subsequent discussions and proofs, thus simplifying the exposition and analysis of the system's dynamics.

\begin{definition}[Truncated augmented state transition function]
\label{def:truncated_augmented_state_transition}
\textnormal{We define the truncated augmented state transition function, denoted as \( \tilde{F}_{\pi}^T:\tilde{\mathcal{X}} \times \mathbb{R}_{\geq 0} \to \tilde{\mathcal{X}}\). This function includes an additional constraint that the initial time \( t \) does not exceed the upper time limit \( T \). The function is specifically defined as
\begin{equation*}
\tilde{F}_{\pi}^T(\tilde{x}, t) = 
\begin{cases} 
    (F_{\pi}(\phi_x(\tilde{x}), t), \phi_t(\tilde{x})+t) & \text{if } \phi_t(\tilde{x}) + t \leq T \\
    (F_{\pi}(\phi_x(\tilde{x}), T-\phi_t(\tilde{x})), T) & \text{if } \phi_t(\tilde{x}) + t > T
\end{cases}.
\end{equation*}
}

\textnormal{
This ensures that the state transition respects the time horizon bound \( T \), truncating the transition at \( T \) when \( \phi_t(\tilde{x})+t \) exceeds it. Given Lemma~\ref{lemma.continuous_state_transition}, which establishes the continuity of \( F_{\pi} \), and considering that the time component is continuous, it follows that \( \tilde{F}_{\pi}^T \) is also continuous.}
\end{definition}

\begin{definition}[Reachable tuple]
\textnormal{Consider the set of initial states \( \mathrm{X}_{\mathrm{init}} \) in the context of a control system with a state transition function \( F_{\pi} \). For a given finite time \( T \), the \emph{reachable tuple} \( \mathcal{R} \) is defined as the collection of all augmented states traversed within the time interval \( [0, T] \) starting from any initial state in \( \mathrm{X}_{\mathrm{init}} \), i.e.,}
\begin{equation*}
\mathcal{R} = \{(F_{\pi}(x_{\mathrm{init}},t),t) \mid 0 \leq t \leq T, x_{\mathrm{init}} \in \mathrm{X}_{\mathrm{init}}\}.
\end{equation*}
\end{definition}

\subsection*{Topological concepts in control system}

In this work, we assume that the state space of the constrained OCP is a subset of Euclidean space. Since Euclidean spaces are topological spaces with the standard topology induced by the Euclidean metric, any subset of Euclidean space, including our state space, can be considered a topological space. Therefore, when we refer to a set of the state space, it is understood to be equipped with the topology inherited from the Euclidean space.

\begin{definition}[Path and loop]
\textnormal{In a set \( \mathrm{X} \), a path is defined as a continuous mapping \( \ell: [0, 1] \to \mathrm{X} \). A path is termed a loop if it satisfies \( \ell(0) = \ell(1) \), indicating that the trajectory forms a closed curve in \( \mathrm{X} \). The reverse of a path \( \ell \) is denoted as \( \ell^{-1} \) and defined by \( \ell^{-1}(s) = \ell(1 - s) \) for all \( s \in [0, 1] \), denoting a path that traverses the same trajectory as \( \ell \) but in the opposite direction.}
\end{definition}

\begin{figure}[!ht]
    \centering
    \includegraphics[width=0.3\textwidth]{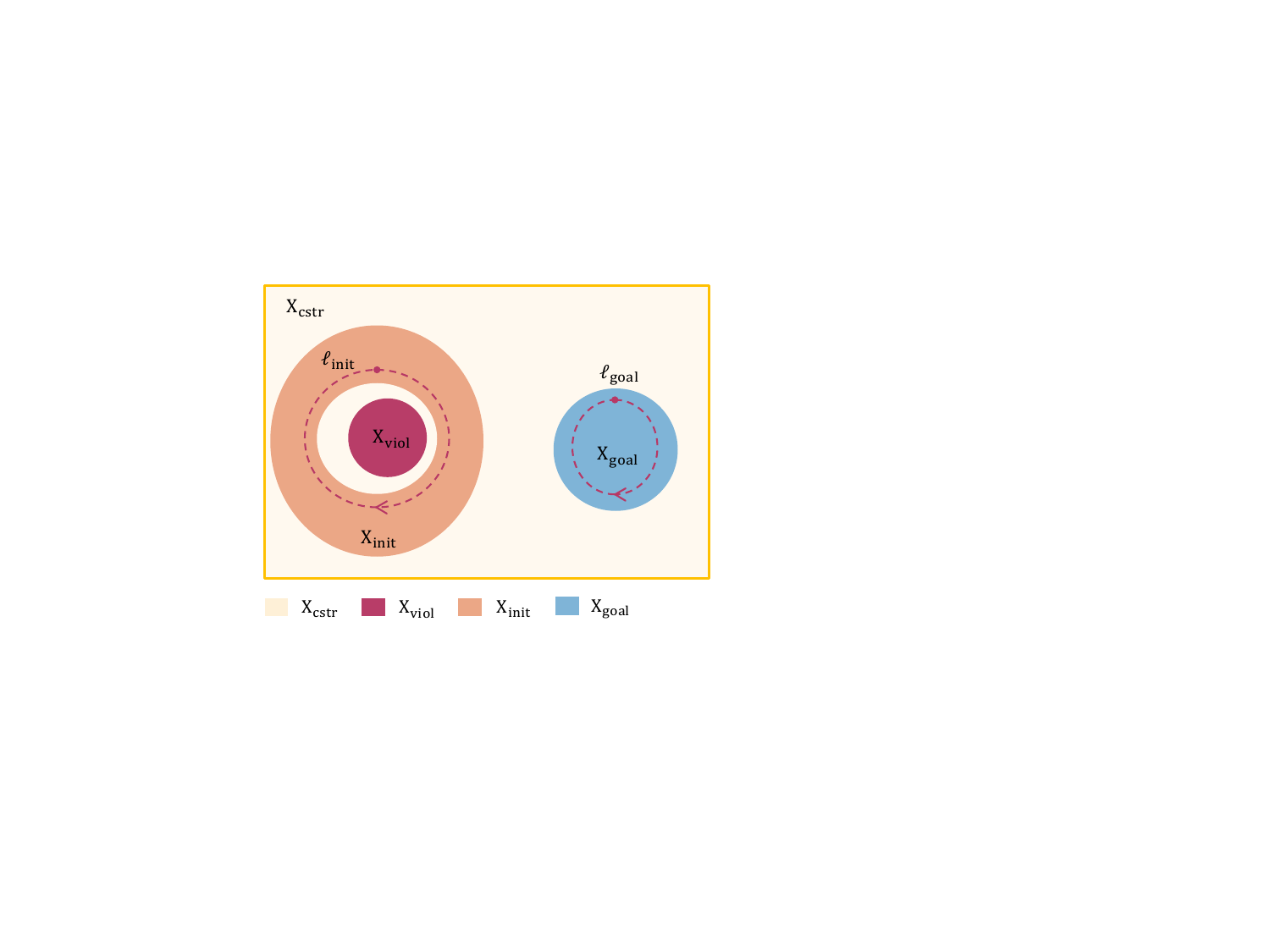}
    \caption{\textbf{Visualization of topological concepts.} This figure illustrates the concepts of contractibility as defined in Definition \ref{def:contractibility}. The space \( \mathrm{X}_{\mathrm{cstr}} \) is non-simply connected due to the presence of \( \mathrm{X}_{\mathrm{viol}} \). The initial state space \( \mathrm{X}_{\mathrm{init}} \) is noncontractible within \( \mathrm{X}_{\mathrm{cstr}} \), as shown by the loop \( \ell_{\mathrm{init}} \), which cannot be continuously contracted to a point within \( \mathrm{X}_{\mathrm{cstr}} \). In contrast, the goal state space \( \mathrm{X}_{\mathrm{goal}} \) is an obstacle-free region, allowing every loop to be continuously contracted to a point, demonstrating its contractibility within \( \mathrm{X}_{\mathrm{cstr}} \).}
    \label{fig:def_of_contractibility}
\end{figure}

The concept of path and loop is important for the concept of ``non-simply connected'' and ``contractibility of a subset''. ``non-simply connected'' is a fundamental concept in topology, referring to a connected space that contains loops that cannot be continuously contracted to a point, usually due to obstacles. The ``contractibility of a subset'' is further detailed in Definition~\ref{def:contractibility}. These concepts, illustrated in Fig.~\ref{fig:def_of_contractibility}, are essential for analyzing the topological properties of the reachable tuple and assessing the feasibility of continuous policies.

\begin{definition}[Contractibility]
\textnormal{Let \( \mathrm{X} \) and \( \mathrm{Y} \) be two sets. \( \mathrm{Y} \) is said to be contractible within \( \mathrm{X} \) if \( \mathrm{Y} \subseteq \mathrm{X} \) and for any loop \( \ell: [0, 1] \to \mathrm{Y} \), there is a point \( c \in \mathrm{X} \) such that \( \ell \) can be continuously deformed into \( c \) within \( \mathrm{X} \). Formally, there exists a continuous mapping \( H: [0, 1] \times [0, T] \to \mathrm{X} \) where \( H(s, 0) = \ell(s) \), and \( H(s, T) = c \) for all \( s \in [0, 1] \). Otherwise, \( \mathrm{Y} \) is said to be noncontractible within \( \mathrm{X} \).}
\label{def:contractibility}
\end{definition}

 For brevity, when we say a set of state or augmented state is contractible or noncontractible, we mean that it is within \(\mathrm{X}_{\mathrm{cstr}}\) or \(\tilde{\mathrm{X}}_{\mathrm{cstr}}\). Based on the lemmas and definitions, the following theorem explores the contractibility of the reachable tuple \(\mathcal{R}\).

\subsection*{Contractibility of the reachable tuple under constraints and continuous dynamics}
\begin{lemma}[Contractibility of the reachable tuple]
\label{lemma.contractibility_of_R}
For a control system with dynamic function \( f \) and policy \( \pi \), the reachable tuple \( \mathcal{R} \) is contractible when the following conditions are satisfied:
\begin{enumerate}
    \item \textbf{Policy continuity}: the policy \( \pi: \mathcal{X} \to \mathcal{U} \) is Lipschitz continuous on \( \mathcal{X} \).
    \item \textbf{Dynamic function continuity}: the dynamic function \( f: \mathcal{X} \times \mathcal{U} \to \mathbb{R}^n \) is Lipschitz continuous on \( \mathcal{X} \times \mathcal{U} \), where \( \mathbb{R}^n \) denotes the space of state derivatives.
    \item \textbf{Safety constraint}: \( F_{\pi}(x(0), t) \in \mathrm{X}_{\mathrm{cstr}} \) for all \( x(0) \in \mathrm{X}_{\mathrm{init}} \) and \( t \in [0, T] \). This implies that \( \mathcal{R} \subseteq \mathrm{\tilde{X}}_{\mathrm{cstr}} \).
    \item \textbf{Finite time reachability}: \( F_{\pi}(x(0), T) \in \mathrm{X}_{\mathrm{goal}} \) for all \( x(0) \in \mathrm{X}_{\mathrm{init}} \).
    \item \textbf{Goal set contractibility}: \( \mathrm{X}_{\mathrm{goal}} \) is contractible.
\end{enumerate}
\end{lemma}

\begin{proof}
To begin with, from Lemma \ref{lemma.continuous_state_transition}, we know that the system's state transition function \( F_{\pi}(x,t) \) is continuous under the conditions of \emph{policy continuity} (condition 1) and \emph{dynamic function continuity} (condition 2). Consequently, the truncated augmented state transition function \( \tilde{F}_{\pi}^T \) is continuous as well.

Let \( \ell_{\mathcal{R}} \) denote any loop in \( \mathcal{R} \), and consider a continuous mapping \( H_{\mathcal{R}}: [0,1] \times [0,T] \to \mathcal{R} \) where \( H_{\mathcal{R}}(s,t) = \tilde{F}_{\pi}^T(\ell_{\mathcal{R}}(s), t) \). This mapping represents the continuous deformation of \( \ell_{\mathcal{R}} \) under the state transition function until \( T \).

Furthermore, since the \emph{safety constraint} (condition 3) ensures \( \mathcal{R} \subseteq \mathrm{\tilde{X}}_{\mathrm{cstr}} \), it follows that for every \( s \in [0,1] \) and \( t \in [0,T] \), we have \( H_{\mathcal{R}}(s,t) \in \mathrm{\tilde{X}}_{\mathrm{cstr}} \).

\(H_{\mathcal{R}}\) leads to the formation of a loop \(\ell_{\mathcal{R},T}(s)= H_{\mathcal{R}}(s,T)\). Projecting \(  \ell_{\mathcal{R},T} \) onto \( \mathcal{X}\) results in another loop, defined as \( \ell_{\mathcal{R},T}^x(s)=\phi_x(H_{\mathcal{R}}(s,T)) = \phi_x(\tilde{F}_{\pi}^T(\ell_{\mathcal{R}}(s), T)) 
 = F_{\pi}(\phi_x(\ell_{\mathcal{R}}(s)), T-\phi_t(\ell_{\mathcal{R}}(s)))\).

Given the \emph{finite time reachability} (condition 4) and the fact that \( \ell_{\mathcal{R}}(s) \in \mathcal{R} \), it follows that \( F_{\pi}(\phi_x(\ell_{\mathcal{R}}(s)), T-\phi_t(\ell_{\mathcal{R}}(s))) \in \mathrm{X}_{\mathrm{goal}} \). Consequently, \(\ell_{\mathcal{R},T}^x\) forms a loop in \(\mathrm{X}_{\mathrm{goal}}\).

According to the \emph{goal set contractibility} (condition 5), a continuous mapping \( H_{c}: [0,1] \times [0,T] \to \mathrm{X}_{\mathrm{cstr}} \) exists such that \(H_{c}(s,0)=\ell_{\mathcal{R},T}^x(s)\), and there exists a constant \( c \in \mathrm{X}_{\mathrm{goal}} \) such that \( H_{c}(s,T) = c \) for all \( s \in [0, 1] \).

Subsequently, we define \(H_{\tilde{c}}(s,t) = (H_{c}(s,t), T)\), which contracts \(\ell_{\mathcal{R},T}(s)\) to a point \(\tilde{c}=(c,T) \in \mathcal{R}\).

Therefore, a continuous mapping \(H_{\mathcal{R},\tilde{c}}: [0,1] \times [0,T] \to \mathrm{\tilde{X}}_{\mathrm{cstr}}\) can be constructed, mapping \(\ell_{\mathcal{R}}\) to a point in \(\mathcal{R}\):
\begin{equation*}
H_{\mathcal{R},\tilde{c}}(s, t) = 
\begin{cases} 
    H_{\mathcal{R}}(s, 2t) & \text{if } t\in[0,T/2] \\
    H_{\tilde{c}}(s,2t-T) & \text{if } t\in(T/2,T]
\end{cases}.
\end{equation*}

In summary, with the fulfillment of conditions 1 to 5, any loop \( \ell_{\mathcal{R}} \) in \( \mathcal{R} \) can be continuously transformed into a point \( \tilde{c} \) in \( \mathrm{\tilde{X}}_{\mathrm{cstr}} \) by \( H_{\mathcal{R},\tilde{c}} \), implying that \( \mathcal{R} \) is contractible.
\end{proof}


\subsection*{Suboptimality and infeasibility of continuous policy in continuous control system}

Lemma~\ref{lemma.contractibility_of_R} demonstrates that for a continuous and feasible policy, the reachable tuple \(\mathcal{R}\) is contractible. This provides a key theoretical foundation for analyzing the performance of continuous policies in constrained OCPs. However, further examination of this theorem suggests that continuous policies may either significantly compromise optimality or lead to violations of constraints.

\begin{theorem}[Suboptimality of continuous policies]
\label{theorem1}
If the optimal solution of an open-loop constrained OCP corresponds to a reachable tuple \( \mathcal{R} \) that cannot be contracted within \( \tilde{\mathrm{X}}_{\mathrm{cstr}} \), then this optimal solution will be discontinuous. Consequently, continuous policies fail to represent this optimal solution of the open-loop constrained OCP.
\end{theorem}

\begin{proof}
We prove this theorem by contradiction. According to Lemma~\ref{lemma.contractibility_of_R}, if a continuous policy \(\pi_{\mathrm{conti}}\) is feasible, then its corresponding reachable tuple \( \mathcal{R} \) is contractible. This implies that if the reachable tuple \( \mathcal{R} \) of \(\pi_{\mathrm{conti}}\) is noncontractible, then \(\pi_{\mathrm{conti}}\) is not feasible. Therefore, for an open-loop OCP whose optimal solution leads to a reachable tuple \( \mathcal{R} \) that is noncontractible, the optimal solution cannot be represented in the form of a continuous policy function. Consequently, continuous policies fail to represent this optimal solution of the open-loop constrained OCP.
\end{proof}

To illustrate the limitations of continuous policies in achieving optimal solutions in certain scenarios, consider the constrained OCP with a 2D state space depicted in Fig.~\ref{fig:theorem1_1}. This figure shows a scenario where the reachable tuple \( \mathcal{R} \) cannot be contracted within \( \tilde{\mathrm{X}}_{\mathrm{cstr}} \). We observe two augmented initial states, \( \tilde{x}_{\mathrm{a}}=(x_{\mathrm{a}}, 0) \) and \( \tilde{x}_{\mathrm{b}}=(x_{\mathrm{b}}, 0) \), and a path \( \ell_{\mathrm{ab}}\) that connects these states and remains within \( \tilde{\mathrm{X}}_{\mathrm{init}} \) for all \(s\in[0,1]\). 

The desired continuous policy, denoted as \(\pi_{\mathrm{des}}\), mandates the vehicle to (1) execute a right detour from \( \tilde{x}_{\mathrm{a}} \) to reach \( \tilde{x}_{\mathrm{a'}} \) and a left detour from \( \tilde{x}_{\mathrm{b}} \) to reach \( \tilde{x}_{\mathrm{b'}} \); (2) ensure collision avoidance with the obstacle and achieve goal set within finite time irrespective of initial state \(x \in \mathrm{X}_\mathrm{init}\). Using Lemma~\ref{lemma.contractibility_of_R}, we will demonstrate that these two conditions present a contradiction for a continuous policy. 

To formalize the trajectories under the desired policy \(\pi_{\mathrm{des}}\), we define paths \(\ell_{\mathrm{aa'}}(s)\) and \(\ell_{\mathrm{bb'}}(s)\). These paths represent the trajectories of the vehicle under the policy \(\pi_{\mathrm{des}}\) from the initial states \( \tilde{x}_{\mathrm{a}} \) and \( \tilde{x}_{\mathrm{b}} \) to their respective goal states \( \tilde{x}_{\mathrm{a'}} \) and \( \tilde{x}_{\mathrm{b'}} \) over a normalized time scale \( s \in [0, 1] \). Specifically, the paths are defined as
\begin{equation*}
\begin{aligned}
    \ell_{\mathrm{aa'}}(s) &= \tilde{F}_{\pi_{\mathrm{des}}}^T(\tilde{x}_{\mathrm{a}}, sT), \\
    \ell_{\mathrm{bb'}}(s) &= \tilde{F}_{\pi_{\mathrm{des}}}^T(\tilde{x}_{\mathrm{b}}, sT).
\end{aligned}
\end{equation*}

\(\ell_{\mathrm{ab}}\) is a path in \(\mathrm{X}_\mathrm{init}\), it starts from \( \tilde{x}_{\mathrm{a}} \), ends at \( \tilde{x}_{\mathrm{b}} \). \(\ell_{\mathrm{a'b'}}\) is a path in \(\mathrm{X}_\mathrm{goal}\), it starts from \( \tilde{x}_{\mathrm{a'}} \), ends at \( \tilde{x}_{\mathrm{b'}} \). Consider the loop \(\ell_{aa'b'ba}\) within \(\mathcal{R}\), which transitions through the sequence \(a \to a' \to b' \to b \to a\):
\begin{equation*}
\ell_{aa'b'ba}(s) = 
\begin{cases}
    \ell_{\mathrm{aa'}}(4s) & \text{for } 0 \leq s \leq \frac{1}{4}, \\
    \ell_{\mathrm{a'b'}}(4s - 1) & \text{for } \frac{1}{4} < s \leq \frac{1}{2}, \\
    \ell_{\mathrm{bb'}}^{-1}(4s - 2) & \text{for } \frac{1}{2} < s \leq \frac{3}{4}, \\
    \ell_{\mathrm{ab}}^{-1}(4s - 3) & \text{for } \frac{3}{4} < s \leq 1.
\end{cases}
\end{equation*}

As depicted in Fig.~\ref{fig:theorem1_1}, the loop \(\ell_{aa'b'ba}\) cannot be continuously contracted to a point within \( \mathrm{\tilde{X}}_{\mathrm{cstr}} \) due to the obstacle, suggesting that the reachable tuple \(\mathcal{R}\) is noncontractible. This noncontractibility contradicts Lemma~\ref{lemma.contractibility_of_R}. Consequently, as shown in Fig.~\ref{fig:theorem1_2}, a feasible continuous policy is limited to only circumventing the obstacle from one side, which allows its reachable tuple to be contractible but results in a lower total reward.

Moreover, when the initial state space \( \mathrm{X}_\mathrm{init} \) is noncontractible, the constrained OCP can become unsolvable, indicating the absence of any feasible continuous policy under such conditions. This concept is further explained in the following theorem, which specifically addresses the conditions leading to the non-existence of solutions in constrained OCPs.

\begin{theorem}[Infeasibility of continuous policies]
\label{theorem2.nonexistence_solution}
Consider a constrained OCP characterized by a Lipschitz continuous dynamic function \( f \) and policy \( \pi \). The problem has no feasible solution if the initial state set \( \mathrm{X}_\mathrm{init} \) is noncontractible, and the goal state set \( \mathrm{X}_\mathrm{goal} \) is contractible. Specifically, this constrained OCP is unsolvable when the following six conditions are simultaneously met:
\begin{enumerate}
    \item \textbf{Policy continuity}: the policy \( \pi: \mathcal{X} \to \mathcal{U} \) is Lipschitz continuous on \( \mathcal{X} \).
    \item \textbf{Dynamic function continuity}: the dynamic function \( f: \mathcal{X} \times \mathcal{U} \to \mathbb{R}^n \) is Lipschitz continuous on \( \mathcal{X} \times \mathcal{U} \), where \( \mathbb{R}^n \) denotes the space of state derivatives.
    \item \textbf{Safety constraint}: \( F_{\pi}(x(0), t) \in \mathrm{X}_{\mathrm{cstr}} \) for all \( x(0) \in \mathrm{X}_{\mathrm{init}} \) and \( t \in [0, T] \). This implies that \( \mathcal{R} \subseteq \mathrm{\tilde{X}}_{\mathrm{cstr}} \).
    \item \textbf{Finite time reachability}: \( F_{\pi}(x(0), T) \in \mathrm{X}_{\mathrm{goal}} \) for all \( x(0) \in \mathrm{X}_{\mathrm{init}} \).
    \item \textbf{Goal set contractibility}: \( \mathrm{X}_{\mathrm{goal}} \) is contractible.
    \item \textbf{Initial state set noncontractibility}: \( \mathrm{X}_{\mathrm{init}} \) is noncontractible.
\end{enumerate}
\end{theorem}

\begin{proof}
As established in Lemma~\ref{lemma.contractibility_of_R}, under conditions 1 to 5, the reachable tuple \( \mathcal{R} \) is contractible. However, by incorporating the \emph{initial state set noncontractibility} (condition 6), we demonstrate that this contractibility does not hold for \( \mathcal{R} \) within \( \mathrm{\tilde{X}}_{\mathrm{cstr}} \). 

According to the definition of contractibility (see Definition \ref{def:contractibility}), condition 6 implies the existence of a loop \( \ell_{\mathrm{init}}: [0, 1] \to \mathrm{X}_{\mathrm{init}} \) such that there is no such continuous mapping \( H: [0, 1] \times [0, T] \to \mathrm{X}_{\mathrm{cstr}} \), where \( H(s, 0) = \ell_{\mathrm{init}}(s) \) and \( H(s, T) = c \) for some \( c \in \mathrm{X}_{\mathrm{cstr}} \) and all \( s \in [0, 1] \). This indicates that \( \ell_{\mathrm{init}} \) cannot be continuously contracted to a point within \( \mathrm{X}_{\mathrm{cstr}} \).

The augmented loop \( \ell_{\mathrm{init}, \mathrm{aug}}(s) = (\ell_{\mathrm{init}}(s), 0) \) forms a loop in \( \mathcal{R} \). To demonstrate the non-existence of a continuous mapping capable of contracting \( \ell_{\mathrm{init}, \mathrm{aug}}\) to a point within \( \mathrm{\tilde{X}}_{\mathrm{cstr}} \), we initially assume that a continuous mapping \( H_{\mathrm{init}, \mathrm{aug}} \) exists. Then, define \( H_{\mathrm{init}}: [0, 1] \times [0, T] \to \mathrm{X}_{\mathrm{cstr}} \) as \( H_{\mathrm{init}}(s, t) = \phi_x(H_{\mathrm{init}, \mathrm{aug}}(s, t)) \). \(H_{\mathrm{init}}\) contracts the loop \( \ell_{\mathrm{init}} \) to a point in \( \mathrm{X}_{\mathrm{cstr}} \). However, this directly conflicts with the noncontractibility of \( \mathrm{X}_{\mathrm{init}} \) as stated in condition 6. Hence, this contradiction demonstrates that \( \mathcal{R} \) is noncontractible under condition 6, since \( \ell_{\mathrm{init}, \mathrm{aug}}\) cannot be contracted to a point within \( \mathrm{\tilde{X}}_{\mathrm{cstr}} \).

In summary, conditions 1 to 5 imply that the reachable tuple \( \mathcal{R} \) is contractible. However, condition 6 leads to the noncontractibility of \( \mathcal{R} \). This contradiction underlines the unsolvability of the constrained OCP under these specific conditions.
\end{proof}

Consequently, continuous policies fail to achieve optimality and feasibility in scenarios with complex constraints, indicating the need for an improved policy structure. The following sections explore Safe RL with bifurcated policies to address the challenges inherent in continuous policies.

\subsection*{Bifurcated policy construction with Gaussian mixture distribution}

In this section, we introduce a construction for bifurcated policy using Gaussian mixture distribution. By selecting the action with the highest probability, the policy's output can abruptly change in response to minor variations in inputs, enabling it to meet state constraints.

\begin{definition}[Bifurcated policy]
\textnormal{A ``bifurcated policy'' is a control policy that exhibits distinct behavioral modes in response to continuous changes in state. This bifurcation is marked by discontinuities at certain critical points $x_c$ in the state space, formalized as
\begin{equation}
\lim_{x \to x_c, x \in \Omega^-} \pi(x) \neq \lim_{x \to x_c, x \in \Omega^+} \pi(x),
\end{equation}
where $\Omega^-$ and $\Omega^+$ are two disjoint regions in the state space that converge to $x_c$ from different directions.
}
\end{definition}

An example of such a critical point is the center line of a road, as illustrated in Fig.~\ref{fig:maneuvering_vehicles}. At these points, the policy behavior exhibits an abrupt change. This characteristic is particularly beneficial in environments with complex safety constraints, where different modes of action are required under varying conditions.

Our approach to constructing a bifurcated policy involves creating a stochastic policy that outputs a Gaussian mixture distribution:
\begin{equation}
    \pi(\cdot| x) = \sum_{i=1}^{k} \pi_{\mathrm{gate}}^i(x; \theta) \mathcal{N}(\mu_i(x; \theta), \sigma_i^2(x; \theta)),
\end{equation}
where \(\pi_{\mathrm{gate}}^i(x; \theta)\) denotes the probability of selecting the \(i\)-th Gaussian component in the mixture distribution, i.e., gate probability. The policy \(\pi(\cdot| x)\) is a mixture of \(k\) Gaussian components, with each component \(\mathcal{N}(\mu_i(x; \theta), \sigma_i^2(x; \theta))\) characterized by its mean \(\mu_i(x; \theta)\) and variance \(\sigma_i^2(x; \theta)\). The parameters \(\theta\) are used to parameterize the mixture distribution. 

To obtain a deterministic action from this mixture distribution, we select the mean of the Gaussian component with the highest gate probability:
\begin{equation}
    u = \mu_{\mathrm{max}}(x; \theta), \quad \text{where }\mathrm{max} = \mathop{\arg\max}\limits_{i} \, \pi_{\mathrm{gate}}^i(x; \theta).
\end{equation}

Through this approach, the stochastic policy with Gaussian mixture distribution is transformed into a bifurcated policy, characterized by discontinuous action outputs that change abruptly depending on the state \(x\). The training process, discussed in subsequent sections, involves learning the parameters \(\theta\) to implement this policy in varying environments effectively.

It is worth noting that the discontinuity in our policy is derived from selecting the maximum gate probability, \(\pi_{\mathrm{gate}}\), instead of learning a very steep continuous function for the mean \(\mu_i(x; \theta)\) of each component. Therefore, we propose the incorporation of spectral normalization \cite{miyato2018spectral} to prevent each component from overfitting on discontinuous segments, thereby promoting the learning of bifurcated policy. This process constrains the Lipschitz constant of the policy network to a predefined limit, denoted by \(L_\pi\).


\subsection*{Bifurcated policy learning}

In this section, we propose MUPO, a safe RL algorithm specifically designed for the efficient learning of bifurcated policy. Built upon the actor-critic architecture, MUPO comprises an actor learning a bifurcated policy \(\pi_{\theta}\) and a critic evaluating the policy. The critic utilizes two distinct action-value distribution functions denoted as \(\mathcal{Z}_{\omega_1}\) and \(\mathcal{Z}_{\omega_2}\), where \(\omega_1\) and \(\omega_2\) represent their parameters respectively.

We utilize the straightforward and effective reward penalty method \cite{thomas2021safe} to handle state constraints. The modified reward function with the constraint \( h(x) \leq 0 \) is defined as follows:

\begin{equation}
\tilde{r}(x, u) = \begin{cases} 
r(x, u) & \text{if } h(x) \leq 0, \\
r(x, u) - C & \text{otherwise}.
\end{cases}
\end{equation}

As long as the penalty \( C \) is sufficiently large, it ensures that the RL algorithm converges to the policy that satisfies the constraints.

\setcounter{subsubsection}{0}
\subsubsection{Policy evaluation}

Our algorithm incorporates action-value distribution from the DSAC algorithm \cite{duan2023dsac} within the critic component, enabling a comprehensive capture of potential future rewards. This integration is particularly crucial given the multimodal nature of our policy function, which often results in a significant variance in state-action values (i.e., Q-values). Compared to methods like SAC \cite{haarnoja2018soft} and PPO \cite{schulman2017proximal}, which utilize deterministic value functions, the action-value distribution in the DSAC algorithm effectively mitigates the overestimation issue in value functions, thereby enhancing policy performance. The loss function for the action-value distribution \(\mathcal{Z}_{\omega}\) is defined as the negative expected log-likelihood of the target value given the current value distribution:
\begin{equation}
J_{\mathcal{Z}}(\omega) = -\underset{\substack{(x,u,r,x') \sim \mathcal{B}, u' \sim \pi_{\bar{\theta}} \\ Z(x',u') \sim \mathcal{Z}_{\bar{\omega}}(\cdot|x',u')}}{\mathbb{E}} \left[ \log P(y_z | \mathcal{Z}_{\omega}(\cdot|x,u)) \right],
\end{equation}
where \(\mathcal{B}\) is the replay buffer containing historical samples. The target value \(y_z\) is computed by:
\begin{equation}
y_z = \tilde{r} + \gamma \left( Z(x', u') - \alpha \log \pi_{\bar{\theta}}(u'|x') \right),
\end{equation}
where \(\tilde{r}\) is the modified immediate reward, \(\gamma\) is the discount factor, and \(\alpha\) is the temperature parameter that balances the exploration-exploitation trade-off.

Given that \( \mathcal{Z}_{\omega} \) is assumed to be Gaussian, it can be expressed as \( \mathcal{Z}_{\omega}(\cdot|x, u) = \mathcal{N}(Q_{\omega}(x, u), \sigma_{\omega}(x, u)^2) \), where \( Q_{\omega}(x, u) \) and \( \sigma_{\omega}(x, u) \) are the mean and standard deviation of the state-action value distribution, respectively.

\subsubsection{Policy improvement}

We adopt an energy-based policy approach in line with algorithms like DSAC and SAC. This approach is particularly effective in balancing exploration and exploitation, especially in high-dimensional action spaces. The energy-based policy \(\pi_{\omega}(u|x)\) is defined as
\begin{equation}
 \pi_{\omega}(u|x) = \frac{\exp\left({\alpha^{-1}}Q_{\omega}(x, u)\right)}{N_{\omega}(x)},
 \label{eq:target_policy}
\end{equation}
where the partition function \(N_{\omega}(x)\) normalizes the distribution. 

Policy optimization aims at aligning the current policy \( \pi_\theta \) as closely as possible with the energy-based policy \( \pi_\omega \). While various projection methods are feasible, utilizing the KL divergence proves to be particularly effective.

The KL divergence is a measure of the difference between two probability distributions, it comes in two forms: forward and reverse. The reverse KL divergence, defined as \( \mathbb{E}_{x \sim \mathcal{B}} \left[ D_\mathrm{KL}(\pi_\theta(\cdot|x) \| \pi_\omega(\cdot|x)) \right] \), is highly efficient for uni-modal policies such as Gaussian distributions. However, in multi-modal distribution scenarios, reverse KL divergence tends to learn only one mode while neglecting others \cite{chan2022greedification}, limiting its effectiveness in capturing the full complexity of such distributions.

On the other hand, the forward KL divergence, expressed as \( \mathbb{E}_{x \sim \mathcal{B}} \left[ D_\mathrm{KL}(\pi_\omega(\cdot|x) \| \pi_\theta(\cdot|x)) \right] \), plays an important role in learning multi-modal distributions. It fosters a wider coverage by the policy, which is essential for exploring diverse behavioral modes. However, the accurate computation of forward KL divergence relies on precise sampling from the energy-based policy \( \pi_\omega \), which is computationally challenging. Consequently, policies learned using forward KL divergence may not achieve high performance.

Our method combines these two forms of KL divergence, harnessing the advantages of both forward and reverse divergences. This combination enables the learned multimodal policies to achieve high performance.

Our policy loss function combines the reverse and forward KL divergences, aiming to balance reward exploitation with modal exploration. The loss function is denoted as
\begin{equation}
\begin{aligned}
& J_{\pi}(\theta) = J_{\mathrm{rev\,KL}} + \lambda J_{\mathrm{fwd\,KL}}, \\
& J_{\mathrm{rev\,KL}} = \mathbb{E}_{x \sim \mathcal{B}} \left[ D_\mathrm{KL}(\pi_\theta(\cdot|x) \| \pi_\omega(\cdot|x)) \right], \\
& J_{\mathrm{fwd\,KL}} = \mathbb{E}_{x \sim \mathcal{B}} \left[D_\mathrm{KL}(\pi_\omega(\cdot|x) \| \pi_\theta(\cdot|x)) \right], \\
\end{aligned}
\end{equation}
where \( \lambda \) modulates the contributions of forward KL divergences. This formulation efficiently exploits high-reward regions while enabling the effective exploration of new potential solutions.

Specifically, the reverse KL divergence is expressed as follows, where terms with no gradient to \(\theta\) are omitted:
\begin{equation}
J_{\mathrm{rev\,KL}} = \mathbb{E}_{x \sim \mathcal{B}, u \sim \pi_{\theta}} \left[ \log \pi_\theta(u|x) - Q_\omega(x, u) \right],
\end{equation}
and the forward KL divergence is given by:
\begin{equation}
J_{\mathrm{fwd\,KL}} = \mathbb{E}_{x \sim \mathcal{B}, u \sim \pi_{\omega}} \left[ -\log \pi_\theta(u|x) \right],
\end{equation}
where we can use gradient-based Langevin sampling \cite{welling2011bayesian} to sample actions from \( \pi_{\omega} \).



The temperature parameter \( \alpha \) is pivotal in moderating the balance between exploration and exploitation behaviors. Drawing on the approach delineated in \cite{haarnoja2018soft}, \( \alpha \) is updated by:
\begin{equation}
\alpha \leftarrow \alpha - \beta_{\alpha} \mathbb{E}_{x \sim \mathcal{B}, u \sim \pi_{\theta}} \left[ -\log \pi_{\theta}(u|x) - \mathcal{H} \right],
\label{eq:update_temperature}
\end{equation}
where \( \mathcal{H} \) is a given constant target entropy. The overall algorithm is detailed in the supplementary materials.

\section*{Code availability}

The custom codes developed for our training environment and the MUPO algorithm are available at the following GitHub repository: \url{https://github.com/THUzouwenjun/MUPO}.

\ifCLASSOPTIONcaptionsoff
  \newpage
\fi
\bibliographystyle{ieeetr}
\bibliography{ref}

\section*{Acknowledgements}
This study is supported by National Key R\&D Program of China with 2022YFB2502901, and NSF China with U20A20334 and 52072213. It is also partially supported by Tsinghua University-Didi Joint Research Center for Future Mobility.

\section*{Author contributions}
W.Z. conceived the idea, conducted the theoretical analysis and experiments, and drafted the paper. Y.L. assisted with real-world experiments. J.D. contributed to theoretical proofs. All authors discussed and interpreted the results, and edited the paper.

\section*{Competing Interests}
The authors declare no competing interests.

\clearpage
\onecolumn
\section*{Algorithmic pseudo-code}

\begin{center}
\begin{minipage}{0.5\linewidth}
\begin{algorithm}[H]
\caption{Multimodal Policy Optimization (MUPO)}
\label{alg:MUPO}
\begin{algorithmic}
\STATE \textbf{Input:} $\theta, \omega_1, \omega_2, \alpha, \mathcal{B},\beta_{\mathcal{Z}}, \beta_{\pi}, \beta_{\alpha}, t$
\STATE \textbf{Initialize:} target networks $\bar{\omega}_1 \leftarrow \omega_1, \bar{\omega}_2 \leftarrow \omega_2, \bar{\theta} \leftarrow \theta$

\FOR{each iteration}
    \FOR{each sampling step}
        \STATE Calculate action $u \sim \pi_{\theta}(u|x)$
        \STATE Get reward $r$ and new state $x'$
        \STATE Store transition $(x, u, r, x')$ in buffer $\mathcal{B}$
    \ENDFOR
    \FOR{each update step}
        \STATE Sample a batch of data from $\mathcal{B}$
        \STATE Update critic using: $\omega_i \leftarrow \omega_i - \beta_{\mathcal{Z}} \nabla_{\omega_i} J_{\mathcal{Z}}(\omega_i)$, \\
        \STATE Update actor using: $\theta \leftarrow \theta - \beta_{\pi} \nabla_{\theta} J_{\pi}(\theta)$\STATE Update temperature $\alpha$\STATE Update the target networks using:
        \STATE \quad$\bar{\omega}_i \leftarrow (1 - t) \cdot \bar{\omega}_i + t \cdot \omega_i$\STATE \quad$\bar{\theta} \leftarrow (1 - t) \cdot \bar{\theta} + t \cdot \theta$\ENDFOR
\ENDFOR
\end{algorithmic}
\end{algorithm}
\end{minipage}
\end{center}

\vspace{1cm}

\section*{Hyperparameters}
\begin{table}[ht]
\small
\centering
\caption{}
\label{table:hyperparameters}
\begin{tabular}{lc}
\toprule
\textbf{Hyperparameter} & \textbf{Value} \\
\midrule
\multicolumn{2}{c}{\textit{Shared}} \\
Discount factor \((\gamma)\) & \(0.99\) \\
Number of hidden layers & 3 \\
Number of hidden neurons & 256 \\
Optimizer & Adam\((\beta_1=0.99, \beta_2=0.999)\) \\
Learning rate \((\beta_{\mathcal{Z}})\) & 1e-3 \(\rightarrow\) 5e-5 \\
Learning rate \((\beta_{V})\) & 1e-3 \(\rightarrow\) 5e-5 \\
Learning rate \((\beta_{\pi})\) & 1e-3 \(\rightarrow\) 5e-5 \\
Learning rate \((\beta_{\alpha})\) & 1e-3 \(\rightarrow\) 5e-5 \\
Activation function & GeLU \\
Batch size & 256 \\
Target entropy & \(-\text{dim}(\mathcal{U})\) \\
Initial temperature & 1.0 \\
\midrule
\multicolumn{2}{c}{\textit{MUPO}} \\
Number of components \((k)\) & 2 \\
forward KL weight \( (\lambda) \) & 1.0 \( \rightarrow\) 0\\
Lipschitz constant \((L_\pi)\) & 1.0 \\

\bottomrule
\end{tabular}
\end{table}

 



\end{document}